\documentclass[11pt]{article}

\usepackage{acl}

\usepackage{times}
\usepackage{latexsym}
\usepackage{booktabs}

\usepackage[utf8]{inputenc}

\usepackage{biditools}
\usepackage{listings}

\usepackage{pdfpages}  

\usepackage[T1]{fontenc}
\usepackage{tcolorbox}
\tcbuselibrary{skins,breakable}

\usepackage{pifont}
\newcommand{\cmark}{\ding{51}}
\newcommand{\xmark}{\ding{55}}
\usepackage{multirow}
\usepackage[utf8]{inputenc}
\usepackage{amsmath}
\usepackage{microtype}

\usepackage{inconsolata}

\usepackage{graphicx}

\title{TalkFa: A Unified Benchmark for Farsi Dialogue Generation and
Understanding}

\author{
\parbox{\textwidth}{
\centering
\large\bfseries
Neda Jamshidi$^{1}$, Kamyar Zeinalipour$^{1}$, Fahimeh Akbari$^{1}$,
Monica Bianchini$^{1}$,\\ Marco Maggini$^{1}$, Marco Gori$^{1}$
}\\[1em]
{\normalfont
$^{1}${University of Siena, DIISM, Via Roma 56, 53100 Siena, Italy}
}
}

\begin{document}
\maketitle
\begin{abstract}
Farsi, spoken by more than 120 million people, lacks a comprehensive benchmark for dialogue generation and understanding. We introduce \textsc{TalkFa}, a unified benchmark comprising three complementary datasets: (1) \textsc{Wiki-FaDial}, 4.2K Wikipedia-grounded dialogues for knowledge-grounded generation; (2) \textsc{DailyDialog-FA}, 6.6K dialogues annotated for dialogue acts and emotions; and (3) \textsc{PlayDial-FA}, 2.1K theatrical dialogues with sentiment labels. While LLMs assist data construction, every dialogue undergoes multi-stage review and revision by native Farsi speakers, and only the final human-approved dialogues are released.
Experiments with six \textsc{LLaMA} and \textsc{Mistral} models show that LoRA substantially improves dialogue generation while requiring only 25--50\% of the training data to recover over 90\% of the final performance gains. Across classification tasks, \textsc{FaBERT} achieves the best dialogue-act performance, \textsc{LoRA-Mistral-7B} performs best on emotion recognition, and \textsc{Mistral-24B} achieves the highest sentiment score.
Human evaluation and independent external validation demonstrate the reliability of the benchmark, while comparisons with GPT-4.1 as an LLM judge reveal that automatic metrics substantially overestimate dialogue quality. Zero-shot evaluation with frontier LLMs further shows that TalkFa remains a challenging benchmark. We will release all datasets, annotation guidelines, code, and checkpoints.
\end{abstract}

\begin{figure}[ht!]
    \centering
    \includegraphics[width=\columnwidth,keepaspectratio]{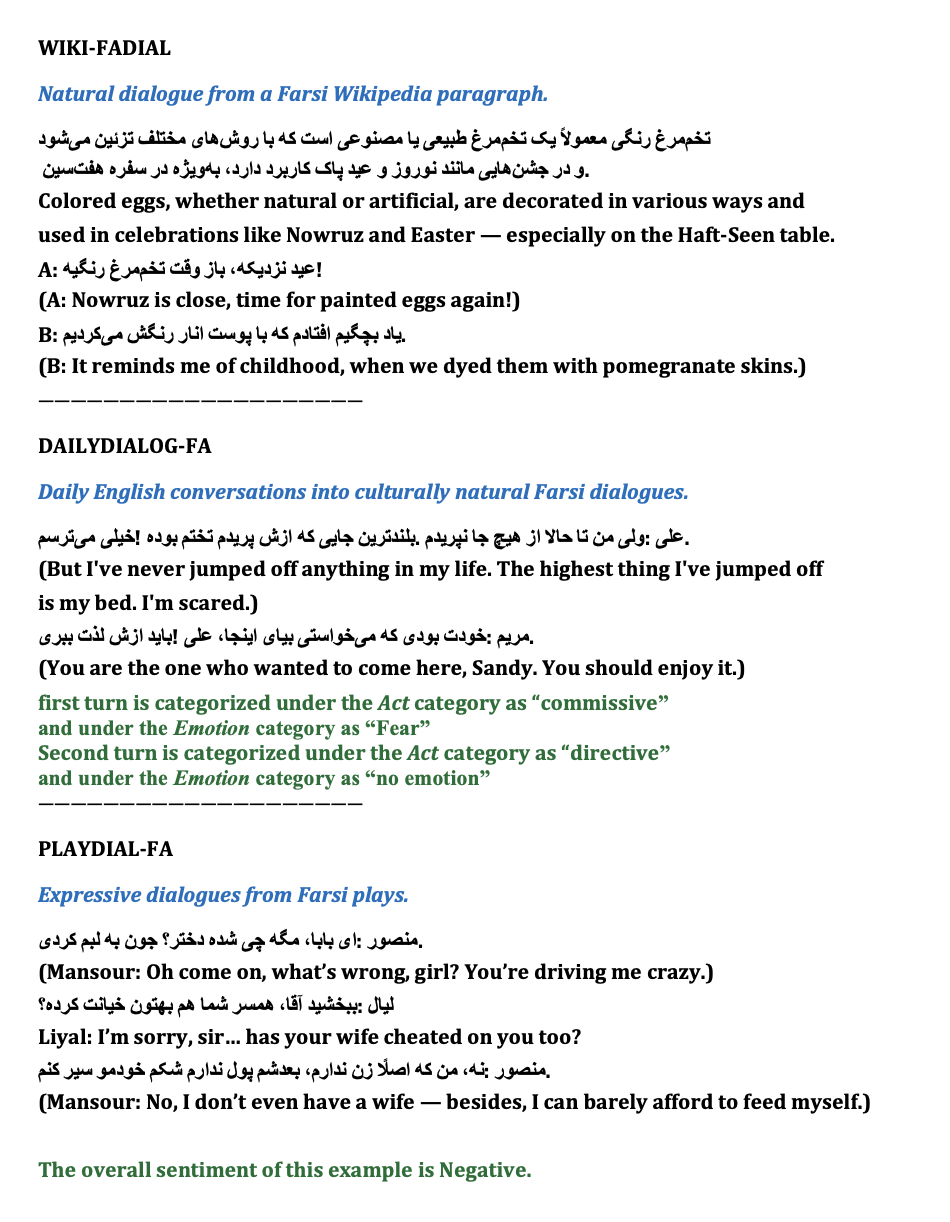}
    \caption{The sample dialogues from three sources
in the TalkFa dataset; full dialogues are provided in Appendix~\ref{app:examples}.}
    \label{fig:Figure1}
\end{figure}

\section{Introduction}
\label{sec:intro}

Progress in conversational AI has been driven by public dialogue benchmarks such as \textsc{DailyDialog}~\cite{li2017dailydialog}, \textsc{EmpatheticDialogues}~\cite{rashkin2018towards}, and \textsc{Topical-Chat}~\cite{gopalakrishnan2023topical}. In contrast, \textbf{Farsi}\footnote{Also called Persian; we use the terms interchangeably.}, spoken by over 120 million people, lacks a unified benchmark for dialogue generation and understanding. Existing resources mainly focus on speech, sentiment, or general NLU (e.g., \textsc{ParsiNLU}~\cite{khashabi2021parsinlu}), leaving conversational modeling largely unexplored. Although multilingual encoders help alleviate data scarcity~\cite{conneau2020unsupervised}, they often fail to capture culturally grounded conversational behavior. Recent work, such as \textsc{NusaDialogue}~\cite{purwarianti2025nusadialogue}, demonstrates the effectiveness of LLM-assisted, human-verified dataset construction, but no comparable benchmark exists for Farsi. The lack of a \emph{public multi-task benchmark} limits reproducible evaluation of Farsi dialogue systems, motivating the following research question:

\begin{tcolorbox}[title=Problem Statement,coltitle=white,colback=pink!15,colframe=pink]
\textbf{How can we build a reliable, culturally adapted benchmark for evaluating Farsi dialogue generation and understanding across multiple dialogue tasks?}
\end{tcolorbox}

To answer this question, we investigate five research questions:

\smallskip
\noindent\textbf{RQ1.} How much does LoRA improve knowledge-grounded Farsi dialogue generation?

\noindent\textbf{RQ2.} Which models perform best for dialogue-act and emotion classification?

\noindent\textbf{RQ3.} How well do the same models perform on theatrical sentiment classification?

\noindent\textbf{RQ4.} How do zero-shot and LoRA-adapted LLMs compare across TalkFa tasks?

\noindent\textbf{RQ5.} How does generation quality scale with training data?

\vspace{0.5em}
\noindent\textbf{Contributions.}
We introduce \textsc{TalkFa}, the first unified benchmark for Farsi dialogue, comprising \textsc{Wiki-FaDial}, \textsc{DailyDialog-FA}, and \textsc{PlayDial-FA}, covering knowledge-grounded dialogue generation, dialogue-act classification, emotion recognition, and sentiment analysis (Figure~\ref{fig:Figure1}). We present a reproducible LLM-assisted, human-curated pipeline through native-speaker review and revision. We establish strong baselines using LoRA-adapted \textsc{LLaMA} and \textsc{Mistral} models together with multilingual and Farsi-specific encoders. Finally, we release all datasets, annotation guidelines, code, LoRA adapters, and fine-tuned checkpoints to support reproducible Farsi dialogue research.

\section{Related Work}
\label{sec:related}

\textbf{Dialogue Benchmarks.}
English dialogue benchmarks such as \textsc{DailyDialog}~\citep{li2017dailydialog}, \textsc{EmpatheticDialogues}~\citep{rashkin2018towards}, \textsc{Topical-Chat}~\citep{gopalakrishnan2023topical}, \textsc{Wizard of Wikipedia}~\citep{dinan2018wizard}, \textsc{PersonaChat}~\citep{zhang2018personalizing}, and \textsc{MultiWOZ}~\citep{budzianowski2018multiwoz} have established benchmarks for open-domain, empathetic, knowledge-grounded, and task-oriented dialogue. Beyond English, Chinese resources include \textsc{KdConv}~\citep{zhou2020kdconv} and \textsc{DuConv}~\citep{wu2019proactive}, while multilingual benchmarks such as \textsc{xDial-Eval}~\citep{zhang2023xdial}, \textsc{MEGA}~\citep{ahuja2023mega}, and \textsc{MTOP}~\citep{li2020mtop} address multilingual dialogue evaluation, broader generative language model evaluation, and task-oriented semantic parsing, respectively. However, no existing benchmark jointly supports Farsi dialogue generation together with multiple dialogue understanding tasks.

\textbf{Knowledge-Grounded Dialogue.}
Knowledge-grounded dialogue has received increasing attention through datasets such as \textsc{Wizard of Wikipedia}~\citep{dinan2018wizard}, \textsc{FaithDial}~\citep{dziri2022faithdial}, and \textsc{CMU-DoG}~\citep{zhou2018dataset}, alongside evaluation benchmarks such as \textsc{BEGIN}~\citep{dziri2022evaluating}. In Farsi, \textsc{Wiki-FaDial} extends this direction using GPT-4o-assisted dialogue generation followed by native-speaker post-editing and automatic and human evaluation.

\textbf{Dialogue Evaluation.}
Automatic metrics including BLEU~\citep{papineni2002bleu}, ROUGE~\citep{lin2004rouge}, and BERTScore~\citep{zhang2019bertscore} remain widely used despite their limitations for open-ended dialogue~\citep{liu2016not}. More recent work introduced learned evaluators such as \textsc{USR}~\citep{mehri2020usr}, \textsc{FED}~\citep{mehri2020unsupervised}, and LLM-as-a-judge methods~\citep{zheng2023judging}. Consistent with these observations, our experiments show that automatic metrics substantially overestimate Farsi dialogue quality relative to human judgments.

\textbf{Synthetic Data Generation and Farsi Resources.}
LLM-assisted dialogue generation has become increasingly common for low-resource languages~\citep{suresh2025diasynth,lee2023making}. \textsc{NusaDialogue}~\citep{purwarianti2025nusadialogue} further demonstrates the use of LLM-assisted generation combined with human validation for dialogue resources in underrepresented languages. While multilingual pretrained models such as mT5~\citep{xue2021mt5} provide broad cross-lingual capabilities, multilingual coverage alone does not ensure culturally grounded conversational modeling. Resources such as \textsc{ArabCulture}~\citep{sadallah2025commonsense} emphasize the importance of culture-aware language technologies. Existing Farsi resources mainly focus on NLU or general LLM evaluation, including \textsc{ParsiNLU}~\citep{khashabi2021parsinlu} and \textsc{FarsEval-PKBETS}~\citep{shamsfard2025farseval}, while \textsc{FaBERT}~\citep{sbunlp_fabert} and \textsc{ParsBERT}~\citep{hooshvarelab_bert_base_parsbert_uncased} are pretrained Farsi language models for downstream tasks. In contrast, to the best of our knowledge, \textsc{TalkFa} is the first unified Farsi benchmark supporting knowledge-grounded dialogue generation, dialogue-act classification, emotion recognition, and sentiment analysis. Table~\ref{tab:related-comparison} compares \textsc{TalkFa} with representative dialogue benchmarks and existing Farsi language resources.

\begin{table}[t]
\centering
\scriptsize
\setlength{\tabcolsep}{2.2pt}
\renewcommand{\arraystretch}{0.95}

\begin{tabular}{@{}llrccccc@{}}
\toprule
\textbf{Dataset} &
\textbf{Language} &
\textbf{Size} &
\textbf{G} &
\textbf{A} &
\textbf{E} &
\textbf{S} &
\textbf{H} \\
\midrule

\textsc{DailyDialog}
& English
& 13K
& \cmark & \cmark & \cmark & \xmark & \cmark \\

\textsc{Wizard of Wikipedia}
& English
& 22K
& \cmark & \xmark & \xmark & \xmark & \cmark \\

\textsc{MultiWOZ}
& English
& 10K
& \cmark & \cmark & \xmark & \xmark & \cmark \\

\textsc{KdConv}
& Chinese
& 4.5K
& \cmark & \xmark & \xmark & \xmark & \cmark \\

\textsc{DuConv}
& Chinese
& 30K
& \cmark & \xmark & \xmark & \xmark & \cmark \\

\textsc{DiaSet}
& Arabic
& 23.2K
& \xmark & \xmark & \xmark & \cmark & \cmark \\

\textsc{NusaDialogue}
& Multilingual
& 30K
& \cmark & \xmark & \xmark & \xmark & \cmark \\

\textsc{MTOP}
& Multilingual
& 100K
& \xmark & \xmark & \xmark & \xmark & \cmark \\

\midrule

\textsc{ParsiNLU}
& Farsi
& 14.5K
& \xmark & \xmark & \xmark & \xmark & \cmark \\

\textsc{FarsEval-PKBETS}
& Farsi
& 4K
& \xmark & \xmark & \xmark & \xmark & \cmark \\

\textbf{\textsc{TalkFa} (ours)}
& \textbf{Farsi}
& \textbf{12.9K}
& \cmark & \cmark & \cmark & \cmark & \cmark \\

\bottomrule
\end{tabular}

\caption{
Comparison with representative English and multilingual dialogue benchmarks together with existing Farsi language resources.
G: dialogue generation; A: dialogue-act classification;
E: emotion classification; S: sentiment classification;
H: human annotation or validation.
Dataset sizes are reported in their original units and may not be directly comparable.
$^\dagger$\textsc{FarsEval-PKBETS} is a general LLM evaluation benchmark rather than a dialogue dataset.
}
\label{tab:related-comparison}
\end{table}

\section{TalkFa Creation}
\label{sec:data}

\begin{table}[t]
\centering
\scriptsize
\setlength{\tabcolsep}{2.5pt}
\renewcommand{\arraystretch}{0.95}

\begin{tabular}{lrrrr}
\toprule
\textbf{Metric} &
\textbf{Parag} &
\textbf{Wiki} &
\textbf{Daily} &
\textbf{Play} \\
\midrule
Tokens(total)         & 1,109,523 & 342,200 & 635,942 & 200,158 \\
Entries(dialogues)         & NA        & 4,184   & 6,599   & 2,070 \\
Turns(total)           & NA        & 25,104  & 52,018  & 16,499 \\
Mean dialogue turns      & NA        & 6.00    & 7.88    & 7.97 \\
Mean tokens dialogue     & 265.18    & 81.79   & 96.37   & 96.69 \\
Median tokens dialogue   & 235       & 81      & 95      & 91 \\
Std.\ deviation (tokens)    & 118.64    & 8.46    & 26.64   & 46.53 \\
 Unique Word types      & 45,157    & 28,448  & 13,624  & 12,025 \\
Unique Lemmas          & 39,477    & 24,394  & 10,664  & 9,646 \\
TTR (words)     & 0.05      & 0.11    & 0.03    & 0.08 \\
TTR (lemmas)    & 0.04      & 0.09    & 0.03    & 0.07 \\
\bottomrule
\end{tabular}

\caption{Corpus statistics for the three TalkFa datasets. ``Parag'' denotes Wiki-FaDial-Parag, ``Wiki'' denotes Wiki-FaDial, ``Daily'' denotes DailyDialog-FA, and ``Play'' denotes PlayDial-FA.}
\label{tab:stats}
\end{table}

Built from three complementary sources, \textsc{TalkFa} covers knowledge-grounded dialogue (\textsc{Wiki-FaDial}, 4.2K six-turn chats), everyday conversation (\textsc{DailyDialog-FA}, 6.6K dialogues with turn-level act and emotion labels), and dramatic dialogue (\textsc{PlayDial-FA}, 2.1K dialogues with dialogue-level sentiment).

\subsection{\textsc{Wiki-FaDial}: Context-Aware Casual Farsi Dialogues}
\label{sec:wikifadial}

We study \emph{context-aware casual conversation generation in Farsi}: given an expository paragraph, the goal is to generate a short, informal dialogue that remains faithful to the source while sounding natural and idiomatic in everyday Farsi. \textsc{Wiki-FaDial} addresses this gap by pairing Farsi encyclopaedic paragraphs with multi-turn conversations in a conversational register.

\paragraph{Construction.}

\textbf{Source discovery.}
We collected pages from ten curated indices of pedagogically valuable Farsi Wikipedia content, including \emph{Featured Articles}, \emph{Vital Articles}, and \emph{Most-Viewed by Topic}~\cite{wikipedia_fa,100_essential_articles,most_viewed,essential_articles_every_wiki_should_have}. This yielded \textbf{8,894} pages spanning mathematics, history, biology, literature, and other core disciplines.

\textbf{Content filtering.}
We retained only the lead paragraph of each article and applied three filters: (i) a \textbf{length filter}, removing paragraphs with fewer than 100 words; (ii) a \textbf{policy filter}, excluding sensitive or disallowed content; and (iii) a \textbf{topicality filter}, retaining self-contained definitions, key facts, or short narratives. This resulted in \textbf{4,184} high-quality source paragraphs.

\textbf{Prompting and generation.}
Each paragraph was converted into a six-turn dialogue using \textsc{GPT-4o}; the full prompt is provided in Appendix~\ref{app:promptwiki}. The prompt required informal, culturally appropriate Farsi, exactly six turns, 1--20 words per turn, and a \texttt{reference} field in every turn to quote or summarize the source paragraph for traceability. The model returned one JSON object per paragraph with the key \texttt{"dialogue"}, whose value is an array of six \{\texttt{speaker}, \texttt{text}, \texttt{reference}\} triples.

\paragraph{Human Verification.}

Two native Farsi annotators\footnote{Both annotators are authors of this paper and performed annotation voluntarily without financial compensation.} with prior NLP experience reviewed and revised every generated dialogue. Both annotators were born and raised in Iran and are familiar with Farsi linguistic conventions and cultural context. In the first stage, one annotator corrected grammatical errors, normalized slang, inserted appropriate discourse markers, and verified the factual consistency of all \texttt{reference} snippets against the source paragraph. In the second stage, the other annotator independently reviewed fluency, internal consistency, and faithfulness to the source text. Because every generated turn includes a \texttt{reference} field, both annotators explicitly verified that each utterance remained grounded in the original paragraph and introduced no unsupported information. Remaining disagreements were resolved through adjudication. Detailed annotation guidelines and editing examples are provided in Appendix~\ref{app:guidelines_wiki} and Appendix~\ref{app:editing_example}.

\paragraph{Dataset Statistics.}

\textbf{\textsc{Wiki-FaDial} features.}
Table~\ref{tab:stats} summarizes corpus statistics for the source paragraphs (\texttt{Wiki-FaDial-Parag}) and the generated dialogues (\texttt{Wiki-FaDial}). The dataset contains \textbf{4,184} dialogues and \textbf{25,104} turns, with a median of \textbf{81} tokens per dialogue (mean \textbf{81.79})\footnote{Using the \href{https://huggingface.co/HooshvareLab/bert-fa-base-uncased}{\texttt{bert-fa}} tokenizer.}. Compared with \textsc{DailyDialog-FA} and \textsc{PlayDial-FA}, it has similar dialogue length but higher lexical diversity (TTR words \textbf{0.11}; lemmas \textbf{0.09}) and relatively low length variance (token std.\ \textbf{8.46}) due to the fixed six-turn design.

\begin{figure}[t]
\centering
\includegraphics[width=\linewidth]{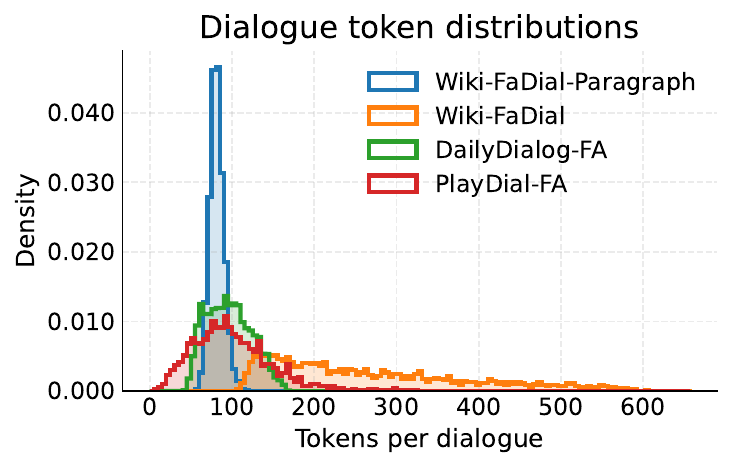}
\caption{Dialogue token distributions across corpora.}
\label{fig:dist}
\end{figure}

\begin{figure*}[t]
\centering
\includegraphics[width=\linewidth]{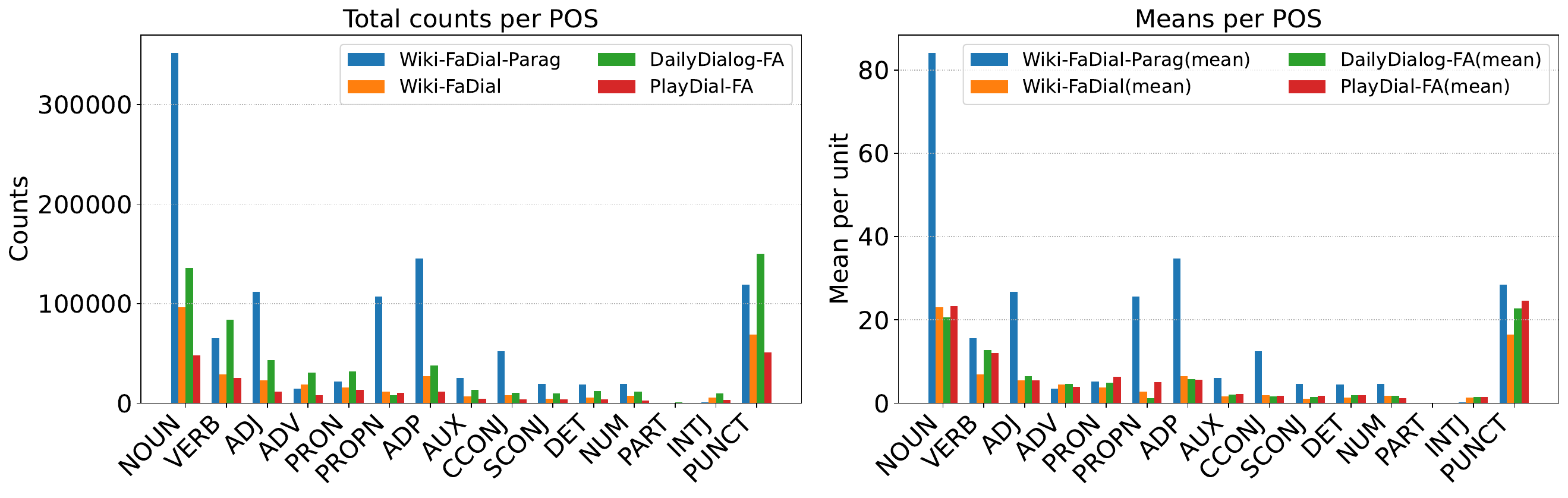}
\caption{POS counts (bars) and per-unit means (lines) across corpora.}
\label{fig:pos}
\end{figure*}

Figure~\ref{fig:dist} shows that \textsc{Wiki-FaDial} dialogues cluster around 80--90 tokens, whereas the source paragraphs are substantially longer. Figure~\ref{fig:pos} shows balanced POS coverage across major grammatical categories, including nouns, verbs, adpositions, pronouns, auxiliaries, and conjunctions\footnote{We performed Farsi POS tagging with Hazm and fell back to Stanza if needed.}.

\textbf{Faithfulness without Copying.}
To measure semantic grounding, we computed embedding similarity between each generated dialogue and its corresponding Wikipedia paragraph using \texttt{bert-fa} sentence embeddings. The average BLEU-4 score is 0.005 and ROUGE-L F$_1$ is 0.10, indicating minimal lexical overlap, while an average embedding cosine similarity of 0.66 confirms that the generated dialogues remain semantically faithful to their source paragraphs.

\enlargethispage{2\baselineskip}
\subsection{\textsc{DailyDialog-FA}: Everyday Farsi Dialogues for Turn-Level Act \& Emotion Classification}
\label{sec:dailydialogfa}

We study \emph{turn-level} dialogue-act and emotion classification. As a culturally localized counterpart to the English \textsc{DailyDialog} corpus, \textsc{DailyDialog-FA} provides Farsi supervision for both tasks while preserving the original annotation scheme.\\
\textbf{Construction.}
\textbf{Label Inventory.}
We preserve the original \textsc{DailyDialog} taxonomy, comprising four dialogue-act categories (\emph{commissive}, \emph{directive}, \emph{inform}, and \emph{question}) and seven emotion categories (\emph{anger}, \emph{disgust}, \emph{fear}, \emph{happiness}, \emph{sadness}, \emph{surprise}, and \emph{no emotion}).

\textbf{Source Corpus and Translation.}
We translate the English \textsc{DailyDialog} corpus~\cite{li2017dailydialog} into Farsi using \textsc{GPT-4o} under constraints that enforce natural spoken Farsi, culturally appropriate references, preservation of dialogue flow, and alignment with Iranian politeness norms. The complete translation prompt is provided in Appendix~\ref{app:model_ids}.

\textbf{Human Verification.}
Two native Farsi annotators reviewed and revised every translated dialogue in two stages. In the first stage, one annotator improved fluency, localized cultural references (e.g., currencies, places, and names), corrected grammatical inconsistencies, and verified semantic fidelity to the English source. In the second stage, the other annotator independently reviewed coherence, clarity, and translation accuracy. Remaining issues were resolved through adjudication, and all dialogues were validated for formatting consistency and correct JSON structure. Since annotators verified translations rather than assigning new labels, inter-annotator agreement is not applicable. Detailed guidelines and editing examples are provided in Appendix~\ref{app:guidelines_translate} and Appendix~\ref{app:editing_example_translate}.

\textbf{Dataset Statistics.}
Table~\ref{tab:stats} summarizes the corpus statistics. \textsc{DailyDialog-FA} contains \textbf{6,599} dialogues, \textbf{52,018} turns, and \textbf{635,942} tokens. Dialogues average \textbf{7.88} turns and \textbf{96.37} tokens (median \textbf{95}; std.\ dev.\ \textbf{26.64}). The dataset exhibits the lexical characteristics of everyday conversational Farsi (types \textbf{13,624}; lemmas \textbf{10,664}; TTR \textbf{0.03}). Figure~\ref{fig:dist} shows that dialogue lengths are concentrated in the short-to-medium range, while Figure~\ref{fig:pos} demonstrates balanced POS coverage across nouns, verbs, pronouns, auxiliaries, and connectives.
\subsection{\textsc{PlayDial-FA}: Dramatic Farsi Dialogues for Dialogue-Level Sentiment Classification}
\label{sec:playdialfa}

We study \emph{dialogue-level sentiment classification} in highly expressive dramatic conversations. Compared with the everyday and knowledge-grounded settings of \textsc{DailyDialog-FA} and \textsc{Wiki-FaDial}, theatrical dialogue exhibits stronger emotional expression, richer rhetorical language, and greater stylistic variation.

\textbf{Construction.}
\textbf{Corpus Harvesting.}
We collected 30 public-domain Farsi plays (1940--2020) from the \textsc{IranNLP Drama Archive}~\cite{takbook_com} and converted them from PDF to text using OCR~\cite{pdf2go_pdf_to_text}. One annotator removed stage directions, formatting artifacts, and OCR errors, while a second annotator verified the cleaned dialogues against the original scripts.

\textbf{Segmentation and Reformulation.}
Using \textsc{GPT-4o}~\cite{openai2024gpt4o}, the plays were segmented into multi-turn dialogues based on \emph{scene continuity}; the complete prompt is provided in Appendix~\ref{app:segmentation-prompt}. To avoid verbatim reproduction while preserving the original style, each dialogue was reformulated by \textsc{GPT-4o} and subsequently reviewed and revised by both annotators for fluency, coherence, and fidelity to the source plays. Detailed guidelines and editing examples are provided in Appendix~\ref{app:play_guidelines} and Appendix~\ref{app:play_example}.

\textbf{Human Verification.}
\textbf{Sentiment Annotation.}
Each dialogue was assigned one of three sentiment labels: \emph{negative}, \emph{neutral}, or \emph{positive}. Initial labels were generated by \textsc{GPT-4o} and independently reviewed by both annotators using the full play context. Disagreements were resolved through discussion until consensus was reached. Inter-annotator agreement reached \textbf{95.89\%} with \textbf{Cohen's $\kappa$ = 0.929}, indicating near-perfect annotation reliability.

\textbf{Dataset Statistics.}
Table~\ref{tab:stats} summarizes the corpus statistics. \textsc{PlayDial-FA} contains \textbf{2,070} dialogues, \textbf{16,499} turns, and \textbf{200,158} tokens. Dialogues average \textbf{7.97} turns and \textbf{96.69} tokens (median \textbf{91}; std.\ dev.\ \textbf{46.53}). Compared with \textsc{DailyDialog-FA}, the dataset exhibits higher lexical diversity (TTR words \textbf{0.08}; lemmas \textbf{0.07}) and greater length variability, reflecting the expressive nature of theatrical dialogue. Figure~\ref{fig:dist} shows a broader dialogue-length distribution, while Figure~\ref{fig:pos} demonstrates balanced POS coverage across verbs, nouns, pronouns, auxiliaries, connectives, and expressive modifiers relevant to sentiment classification.
\subsection{Human Revision Analysis and Dataset Quality}
\enlargethispage{\baselineskip}
\label{sec:quality-verification}
All three datasets underwent the human verification procedures described earlier. To quantify human intervention, we computed normalized character-level edit distances between the initial GPT-4o outputs and the final human-approved dialogues using Python's \texttt{SequenceMatcher}. Annotators were free to revise dialogue structure, wording, speaker turns, pragmatic expressions, and cultural references rather than merely correcting grammatical errors.

\textbf{\textsc{Wiki-FaDial}.}
All 4,184 GPT-4o-generated dialogues were reviewed by native Farsi speakers following Appendix~\ref{app:guidelines_wiki}. Overall, 2,661 dialogues (63.0\%) required manual revision before approval, indicating substantial human intervention.

\textbf{\textsc{DailyDialog-FA}.}
Of the 6,599 translated dialogues, 1,123 (17.0\%) required revision. At the turn level, 1,485 of 52,018 turns (2.9\%) were modified, with a mean edit rate of 12.7\% (median 9.5\%). Most edits improved fluency, phrasing, or cultural adaptation rather than correcting translation errors.

\textbf{\textsc{PlayDial-FA}.}
The 2,070 reformulated dialogues underwent two review rounds. In the first, 395 dialogues (19.1\%) were revised (mean edit rate 8.2\%); in the second, only 28 (1.4\%) required additional changes (mean edit rate 10.4\%), indicating convergence toward stable, human-approved quality.
Table~\ref{tab:edit-stats} summarizes the post-editing statistics. The high proportion of manually revised dialogues, particularly in \textsc{Wiki-FaDial}, confirms that TalkFa is an LLM-assisted, human-curated benchmark. All released dialogues were reviewed and approved by native Farsi speakers.
\begin{table}[t]
\centering
\scriptsize
\setlength{\tabcolsep}{3pt}
\renewcommand{\arraystretch}{0.95}

\begin{tabular}{lccc}
\toprule
\textbf{Dataset} & \textbf{Items} & \textbf{Edited (\%)} & \textbf{Edit Rate} \\
\midrule
Wiki-FaDial    & 4,184 Dialogues.  & \multicolumn{2}{c}{In-place editing} \\
DailyDialog-FA & 52,018 Turns.  & 2.9\%  & 12.7\% \\
PlayDial-FA R1 & 2,070 Dialogues.  & 19.1\% & 8.2\% \\
PlayDial-FA R2 & 2,070 Dialogues.  & 1.4\%  & 10.4\% \\
\bottomrule
\end{tabular}

\caption{Human post-editing statistics. Edited is the percentage of modified items, and Edit Rate is the mean normalized character-level edit distance over modified items only.}
\label{tab:edit-stats}
\end{table}

\subsubsection{Independent External Validation}

To assess dataset quality and annotation reproducibility, an independent native Farsi speaker evaluated 300 stratified random samples (100 per annotation task). \textsc{Wiki-FaDial} achieved 2.8/3.00 factual grounding, 3.65/4.00 naturalness, and 2.70/3.00 cultural appropriateness. \textsc{DailyDialog-FA} achieved 3.00/3.00 meaning preservation, 3.90/4.00 naturalness, and 2.8/3.00 cultural localization, while \textsc{PlayDial-FA} obtained 3.50/4.00 for dialogue naturalness. Agreement with the released annotations reached 85\% ($\kappa=0.800$) for dialogue acts, 89\% ($\kappa=0.872$) for emotions, and 87\% ($\kappa=0.805$) for sentiment, indicating substantial to almost perfect agreement. Full details are provided in Appendix~\ref{app:external_validation}.
\section{Experiments}
\label{sec:experiments}
\begin{table}[t]
\centering
\scriptsize
\setlength{\tabcolsep}{2pt}
\renewcommand{\arraystretch}{0.88}

\begin{tabular}{lrrrrrrrr}
\toprule
\textbf{Model}
& \multicolumn{2}{c}{\textbf{ROUGE-L}$\uparrow$}
& \multicolumn{2}{c}{\textbf{BLEU-4}$\uparrow$}
& \multicolumn{2}{c}{\textbf{BERTScore$_F$$\uparrow$}}
& \multicolumn{2}{c}{\textbf{BERTCos}$\uparrow$} \\
\cmidrule(lr){2-3}\cmidrule(lr){4-5}\cmidrule(lr){6-7}\cmidrule(lr){8-9}
& Base & +LoRA & Base & +LoRA & Base & +LoRA & Base & +LoRA \\
\midrule
Llama-3.2-1B & 0.449 & 0.650 & 0.322 & 0.572 & 0.865 & 0.918 & 0.824 & 0.959 \\
Llama-3.2-3B & 0.353 & 0.660 & 0.229 & 0.582 & 0.836 & 0.922 & 0.714 & 0.959 \\
Mistral-7B   & 0.601 & 0.683 & 0.499 & 0.614 & 0.899 & 0.927 & 0.895 & 0.969 \\
Llama-3.1-8B & 0.655 & 0.668 & 0.578 & 0.596 & 0.912 & 0.925 & 0.944 & 0.962 \\
Mistral-Nemo & 0.613 & \textbf{0.688} & 0.515 & 0.614 & 0.894 & 0.928 & 0.893 & \textbf{0.973} \\
Mistral-24B  & 0.546 & 0.687 & 0.434 & \textbf{0.617} & 0.901 & \textbf{0.930} & 0.919 & 0.972 \\
\bottomrule
\end{tabular}

\caption{Base vs.\ +LoRA performance on \textsc{Wiki-FaDial} (\textit{test}). Bold marks the best score for each metric.}
\label{tab:wikifadial-auto}
\end{table}

\begin{table*}[t]
\centering
\scriptsize
\begin{tabular}{lcc|cc|cc}
\toprule
& \multicolumn{2}{c|}{\textbf{Act}} & \multicolumn{2}{c|}{\textbf{Emotion}} & \multicolumn{2}{c}{\textbf{Sentiment}} \\
\cmidrule(r){2-3}\cmidrule(lr){4-5}\cmidrule(l){6-7}
\textbf{Model} & \textbf{Base} & \textbf{+LoRA} & \textbf{Base} & \textbf{+LoRA} & \textbf{Base} & \textbf{+LoRA} \\
& (P | R | F$_1$) & (P | R | F$_1$) & (P | R | F$_1$) & (P | R | F$_1$) & (P | R | F$_1$) & (P | R | F$_1$) \\
\midrule
Llama-1B &
0.19|0.23|0.21 & 0.66|0.63|\textbf{0.57} &
0.15|0.14|0.13 & 0.24|0.35|\textbf{0.24} &
0.16|0.33|0.21 & 0.33|0.34|\textbf{0.23} \\

Llama-3B &
0.46|0.41|0.39 & 0.63|0.60|\textbf{0.56} &
0.24|0.24|0.19 & 0.29|0.34|\textbf{0.31} &
0.33|0.45|0.38 & 0.58|0.52|\textbf{0.50} \\

Mistral-7B &
0.53|0.23|0.30 & 0.71|0.69|\textbf{0.70} &
0.24|0.21|0.18 & 0.51|0.33|\textbf{0.38} &
0.56|0.08|0.13 & 0.60|0.58|\textbf{0.57} \\

Llama-8B &
0.51|0.48|0.49 & 0.71|0.70|\textbf{0.70} &
0.20|0.22|0.13 & 0.36|0.29|\textbf{0.31} &
0.82|0.18|0.22 & 0.64|0.55|\textbf{0.55} \\

Mistral-12B &
0.52|0.51|0.49 & 0.70|0.68|\textbf{0.68} &
0.26|0.22|0.17 & 0.38|0.29|\textbf{0.32} &
0.58|0.54|0.53 & 0.63|0.57|\textbf{0.57} \\

Mistral-24B &
0.58|0.47|0.48 & 0.73|0.71|\textbf{0.72} &
0.28|0.36|0.27 & 0.52|0.33|\textbf{0.37} &
0.63|0.52|0.53 & 0.65|0.60|\textbf{0.62} \\
\bottomrule
\end{tabular}
\caption{Macro-averaged Precision | Recall | F$_1$ for \textbf{Act}, \textbf{Emotion}, and \textbf{Sentiment}, computed from the model outputs. In each model–category pair, the higher F$_1$ is in bold. Higher is better.}
\label{tab:llm_results_corrected}
\end{table*}

\begin{table*}[t]
\centering
\scriptsize
\begin{tabular}{lcc|cc|cc}
\toprule
& \multicolumn{2}{c|}{\textbf{Act}} & \multicolumn{2}{c|}{\textbf{Emotion}} & \multicolumn{2}{c}{\textbf{Sentiment}} \\
\cmidrule(r){2-3}\cmidrule(lr){4-5}\cmidrule(l){6-7}
\textbf{Model} & \textbf{MLP} & \textbf{Full}
& \textbf{MLP} & \textbf{Full}
& \textbf{MLP} & \textbf{Full} \\
& (P | R | F$_1$) & (P | R | F$_1$)
& (P | R | F$_1$) & (P | R | F$_1$)
& (P | R | F$_1$) & (P | R | F$_1$) \\
\midrule
e5-small-100M &
0.65|0.69|0.66 & 0.71|0.75|\textbf{0.72} &
0.20|0.43|0.21 & 0.26|0.51|\textbf{0.30} &
0.47|0.47|\textbf{0.47} & 0.31|0.33|0.31 \\

fabert-100M &
0.67|0.72|0.68 & 0.73|0.77|\textbf{0.75} &
0.25|0.55|0.28 & 0.29|0.52|\textbf{0.33} &
0.49|0.50|\textbf{0.49} & 0.16|0.33|0.21 \\

roberta-base-100M &
0.51|0.54|0.51 & 0.66|0.69|\textbf{0.67} &
0.02|0.14|0.03 & 0.12|0.14|\textbf{0.13} &
0.42|0.42|0.41 & 0.51|0.52|\textbf{0.51} \\

parsbert-110M &
0.68|0.71|0.69 & 0.72|0.76|\textbf{0.73} &
0.24|0.48|0.26 & 0.30|0.47|\textbf{0.33} &
0.52|0.53|\textbf{0.49} & 0.16|0.33|0.21 \\

bert-base-multicased-200M &
0.59|0.61|0.59 & 0.71|0.75|\textbf{0.72} &
0.20|0.38|0.20 & 0.26|0.55|\textbf{0.29} &
0.43|0.46|0.41 & 0.54|0.55|\textbf{0.54} \\

e5-base-300M &
0.67|0.71|0.68 & 0.72|0.76|\textbf{0.73} &
0.24|0.48|0.26 & 0.31|0.59|\textbf{0.36} &
0.47|0.48|0.47 & 0.56|0.57|\textbf{0.56} \\

xlm-roberta-base-300M &
0.54|0.56|0.54 & 0.72|0.76|\textbf{0.73} &
0.12|0.14|0.13 & 0.30|0.60|\textbf{0.34} &
0.50|0.50|\textbf{0.50} & 0.31|0.38|0.32 \\

roberta-large-400M &
0.51|0.53|0.52 & 0.65|0.69|\textbf{0.66} &
0.15|0.26|0.09 & 0.12|0.14|\textbf{0.13} &
0.46|0.44|\textbf{0.44} & 0.16|0.33|0.21 \\

e5-large-600M &
0.69|0.73|0.70 & 0.72|0.75|\textbf{0.73} &
0.25|0.50|0.26 & 0.31|0.61|\textbf{0.36} &
0.59|0.58|\textbf{0.58} & 0.51|0.51|0.50 \\

xlm-roberta-large-600M &
0.53|0.56|0.54 & 0.72|0.76|\textbf{0.73} &
0.12|0.14|\textbf{0.13} & 0.02|0.14|0.03 &
0.57|0.52|\textbf{0.52} & 0.40|0.50|0.45 \\
\bottomrule
\end{tabular}
\caption{macro-averaged Precision | Recall | F$_1$ for \textbf{Act}, \textbf{Emotion}, and \textbf{Sentiment}, under two fine-tuning regimes (MLP vs.\ Full). In each triplet the higher F$_1$ is in bold.}
\label{tab:corrected_results}
\end{table*}

This section evaluates \textbf{TalkFa} on two tasks: \emph{dialogue generation} and \emph{utterance-level classification}.

\noindent\textbf{Model families.}
We evaluate two model groups.
\emph{(A) Instruction-tuned LLMs} (generation and zero-shot classification):
\textsc{Llama-3.2-1B}~\cite{meta_llama3.2_1B_instruct},
\textsc{Llama-3.2-3B}~\cite{meta_llama3.2_3B_instruct},
\textsc{Llama-3.1-8B}~\cite{meta_llama3.1_8B_instruct},
\textsc{Mistral-7B}~\cite{mistral7b_instruct_v0.3},
\textsc{Mistral-Nemo-12B}~\cite{mistral_nemo_instruct_2407},
and \textsc{Mistral-Small-24B}~\cite{mistral_small_24B_instruct_2501}, spanning 1B--24B parameters.
\emph{(B) Sentence encoders} (supervised classification):
\textsc{Multilingual-E5} Small/Base/Large~\cite{intfloat_multilingual_e5_small,intfloat_multilingual_e5_base,intfloat_multilingual_e5_large},
\textsc{ParsBERT}~\cite{hooshvarelab_bert_base_parsbert_uncased},
\textsc{FaBERT}~\cite{sbunlp_fabert},
mBERT~\cite{google_bert_base_multilingual_cased},
RoBERTa-Base/Large~\cite{facebookai_roberta_base,facebookai_roberta_large},
and XLM-R Base/Large~\cite{facebookai_xlm-roberta_base,facebookai_xlm-roberta_large}. This setup enables comparison between multilingual and Farsi-specific pretraining across model scales.
\noindent\textbf{Training regimes.}
LLMs are evaluated in zero-shot and LoRA-adapted settings~\citep{hu2022lora}. For generation, one adapter is trained on \textsc{Wiki-FaDial}; for classification, separate adapters are trained on \textsc{DailyDialog-FA} and \textsc{PlayDial-FA}. Only Lora adapters and task heads are updated.
Encoders are evaluated under two regimes: (i) frozen backbone with an MLP classifier head, and (ii) full fine-tuning of encoder and classifier jointly.

\noindent\textbf{Training details.}
Experiments are conducted on 3$\times$NVIDIA RTX A6000 GPUs (48GB) using bf16 precision, FlashAttention, and gradient checkpointing.
\emph{LLM LoRA fine-tuning:}
3 epochs; AdamW; cosine scheduler; learning rate $1\times10^{-4}$; weight decay $1\times10^{-4}$; batch size 4 with gradient accumulation 2; LoRA rank 128, $\alpha$ 256, dropout 0.1.
\emph{Encoder classification:}
Frozen-backbone models use a 4-layer MLP head (512$\rightarrow$256$\rightarrow$128$\rightarrow$64) with NLLLoss for 100 epochs (batch size 64; LR $1\times10^{-5}$). Full fine-tuning updates encoder and classifier jointly for 3 epochs (batch size 32; LR $2\times10^{-5}$).

\noindent\textbf{Evaluation.}
For classification, we report accuracy, macro-precision, macro-recall, and macro-F$_1$. For dialogue generation, we report BLEU-4, ROUGE-L, BERTScore-F using \texttt{xlm-roberta-large}, and BERTCos using \texttt{bert-fa}~\cite{hooshvarelab_bert_fa_base_uncased} sentence embeddings.
\enlargethispage{\baselineskip}
\enlargethispage{2\baselineskip}
\subsection{\textsc{Wiki-FaDial}: Dialogue Generation (RQ1)}
\label{sec:exp-wikifadial}
 We evaluate six instruction-tuned LLMs in both \textbf{Base} and \textbf{+LoRA} settings on \textsc{Wiki-FaDial}\footnote{All experiments use a 90/5/5 train/validation/test split.} using the six-turn generation schema from Sec.~\ref{sec:wikifadial}.\\
\textbf{Overall performance.}
LoRA consistently improves performance across all metrics and model sizes (Table~\ref{tab:wikifadial-auto}). Gains are largest for smaller models: \textsc{Llama-3.2-1B} and \textsc{Llama-3.2-3B} nearly double their ROUGE-L scores after fine-tuning, while larger models (\textsc{Mistral-Nemo}, \textsc{Mistral-Small-24B}) show smaller but stable improvements. Notably, a LoRA-adapted 7B model rivals or exceeds a 24B Base model, suggesting that \emph{adaptation matters more than scale} for this task. However, automatic metrics provide only a partial picture: models achieving BERTCos scores above 0.97 still obtain only moderate human ratings.

\subsubsection{Beyond Automatic Metrics: Human and LLM-Based Evaluation}
\label{sec:beyond-auto}

We complement automatic metrics with human evaluation, GPT-4.1-based ~\cite{openai2025gpt41} judging, multilingual semantic scoring, and zero-shot evaluation using frontier LLMs.\\
\textbf{Human evaluation.}
We evaluate 7,200 generated dialogues across all six LLM families, including both Base and +LoRA variants. Each annotator independently rates 3,600 dialogues on a five-point scale from highly natural and coherent (A=5) to incoherent or irrelevant (E=1). One annotator additionally re-annotates a shared subset of 1,200 dialogues for agreement estimation. Detailed guidelines and examples are provided in Appendix~\ref{app:human_eval_guidelines}. Table~\ref{tab:human-eval} reports the results.
\begin{table}[t]
\centering
\scriptsize
\setlength{\tabcolsep}{3pt}
\renewcommand{\arraystretch}{0.9}
\begin{tabular}{lcc|lcc}
\toprule
Model & Base & +LoRA & Model & Base & +LoRA \\
\midrule
Mistral-Small-24B & 1.28 & \textbf{2.74} &
Llama3.1-8B & 1.61 & 1.91 \\

Mistral-Nemo-12B & 1.03 & 2.15 &
Llama3.2-1B & 1.06 & 1.81 \\

Llama3.2-3B & 1.12 & 2.03 &
Mistral-7B & 1.04 & 1.70 \\
\bottomrule
\end{tabular}
\caption{Human evaluation (1--5) on \textsc{Wiki-FaDial}.}
\label{tab:human-eval}
\end{table}
The strongest model (\textsc{Mistral-Small-24B} +LoRA) achieves only \textbf{2.74/5}, indicating that the benchmark remains far from saturated. Re-annotation yields 84.96\% raw agreement and Cohen's $\kappa = 0.759$~\cite{landis1977measurement}, indicating substantial agreement.\\
\textbf{Zero-shot frontier LLMs (RQ4).}
 To assess whether frontier commercial models trivially solve TalkFa tasks, we evaluate GPT-4.1, GPT-4.1-Nano, and DeepSeek-V4-Flash ~\cite{deepseek_v4_flash}in a zero-shot setting (Table~\ref{tab:llm-zeroshot}).
\begin{table}[t]
\centering
\scriptsize
\setlength{\tabcolsep}{3pt}
\renewcommand{\arraystretch}{0.9}
\begin{tabular}{lccc}
\toprule
Task & Model & Acc. & Macro-F$_1$ \\
\midrule
\multirow{3}{*}{Act}
 & GPT-4.1      & \textbf{0.737} & \textbf{0.629} \\
 & GPT-4.1-Nano    & 0.674 & 0.589 \\
 & DeepSeek-V4-Flash      & 0.732 & 0.611 \\
\midrule
\multirow{3}{*}{Emotion}
 & GPT-4.1      & 0.726 & 0.340 \\
 & GPT-4.1-Nano    & 0.722 & 0.330 \\
 & DeepSeek-V4-Flash      & \textbf{0.759} & \textbf{0.353} \\
\midrule
\multirow{3}{*}{Sentiment}
 & GPT-4.1      & 0.664 & 0.613 \\
 & GPT-4.1-Nano    & \textbf{0.712} & \textbf{0.683} \\
 & DeepSeek-V4-Flash      & 0.596 & 0.532 \\
\bottomrule
\end{tabular}
\caption{Zero-shot results on TalkFa.}
\label{tab:llm-zeroshot}
\end{table}
GPT-4.1 achieves 73.7\% accuracy on dialogue acts and 66.4\% on sentiment classification, but emotion classification remains difficult for all models (macro-F\textsubscript{1} $\leq 0.35$), partly due to strong class imbalance.\\
\textbf{LLM-as-a-judge.}
We also use GPT-4.1 (temperature~=~0) to rate the same 7,200 generated dialogues on the A--E scale. Table~\ref{tab:llm-judge} compares GPT-4.1 ratings against human annotations.
\begin{table}[t]
\centering
\scriptsize
\setlength{\tabcolsep}{3pt}
\renewcommand{\arraystretch}{0.9}
\begin{tabular}{lccc}
\toprule
Model & $\rho$ & $\pm$1 & $\kappa_{\rm lin}$ \\
\midrule
Mistral-7B      & \textbf{0.537} & \textbf{92.4} & \textbf{0.387} \\
Llama3.2-1B   & 0.420 & 84.8 & 0.313 \\
Llama3.2-3B   & 0.412 & 84.2 & 0.309 \\
Mistral-Small   & 0.486 & 75.4 & 0.307 \\
Llama3.1-8B   & 0.381 & 73.5 & 0.188 \\
Mistral-Nemo    & 0.334 & 68.0 & 0.193 \\
\midrule
Average. & \textit{0.428} & \textit{79.7} & \textit{0.283} \\
\bottomrule
\end{tabular}
\caption{GPT-4.1 judge vs. human annotators. $\rho$: Spearman ($p<0.001$); $\pm$1: \% within one rating; $\kappa_{\rm lin}$: weighted Cohen's $\kappa$.}
\label{tab:llm-judge}
\end{table}
GPT-4.1 shows moderate agreement with human preferences but is systematically more lenient, assigning on average +0.45 higher grades.\\
\textbf{Data-efficiency ablation (RQ5).}
 LoRA fine-tuning is repeated using \(\{25\%,50\%,75\%,100\%\}\) of \textsc{Wiki-FaDial}. Performance improves rapidly with the first 25\% of training data, and 25--50\% already recovers over 90\% of final gains. Full results are provided in Appendix~\ref{app:data-efficiency}.\\
\textbf{The Automatic--Human Gap.}
Automatic metrics substantially overestimate dialogue quality: models exceeding 0.93 BERTScore and 0.97 BERTCos still receive only moderate human ratings (2.74/5). This suggests that reference-based metrics alone are insufficient for evaluating open-ended Farsi dialogue. Detailed comparisons are provided in Appendix~\ref{app:metric-gap}.
\enlargethispage{\baselineskip}
\enlargethispage{2\baselineskip}
\subsection{\textsc{DailyDialog-FA}: Turn-Level Act \& Emotion Classification (RQ2) }
\label{sec:exp-dailydialogfa-combined}

 We compare six instruction-tuned decoder LMs (Base vs.\ +LoRA) and ten encoders (MLP vs.\ Full) for \textbf{Dialogue-Act} and \textbf{Emotion} classification (Tables~\ref{tab:llm_results_corrected}--\ref{tab:corrected_results}).\\
\textbf{Dialogue acts.}
Farsi-specific encoders perform best overall: \textsc{FaBERT} achieves the highest act macro-F\textsubscript{1} (0.75), outperforming larger multilingual models. LoRA-adapted LLMs narrow the gap---\textsc{Mistral-24B} reaches 0.72---but do not surpass smaller Farsi-pretrained encoders, suggesting that language-specific pretraining is especially beneficial for structured classification in morphologically rich languages such as Farsi.\\
\textbf{Emotions.}
Emotion classification remains the most difficult TalkFa task, with no model exceeding 0.38 macro-F\textsubscript{1}. This likely reflects both severe class imbalance (the dominant ``no emotion'' class) and cultural mismatch between English emotion categories and Farsi pragmatic conventions such as \emph{ta'arof}. Interestingly, multilingual similarity models (\textsc{E5-Base/Large}) slightly outperform Farsi-specific encoders, suggesting that cross-lingual transfer may help low-frequency emotion categories.\\
\textbf{Practical recommendation.}
Adaptation is the main driver of performance: LoRA consistently improves decoder LMs, while full fine-tuning outperforms frozen encoders. Notably, LoRA-adapted 7B models approach the performance of 12B--24B variants, making \textsc{Mistral-7B} a strong trade-off between quality and efficiency.
\enlargethispage{\baselineskip}
\subsection{\textsc{PlayDial-FA}: Sentiment Classification (RQ3) }
\label{sec:exp-playdialfa-sentiment}
 We evaluate dialogue-level sentiment classification (negative / neutral / positive) on expressive theatrical dialogues (Tables~\ref{tab:llm_results_corrected}--\ref{tab:corrected_results}).\\
\textbf{LoRA unlocks LLM potential.}
Base LLMs struggle on literary dialogue, typically showing high precision but very low recall by overpredicting the majority class. LoRA substantially improves performance: \textsc{Mistral-24B} achieves the best result (0.62 macro-F\textsubscript{1}) by recovering minority-class recall.\\
\textbf{Encoders.}
Full fine-tuning benefits some multilingual backbones (\textsc{E5-Base}: 0.56; \textsc{mBERT}: 0.54) but harms others. Notably, Farsi-specific encoders (\textsc{FaBERT}, \textsc{ParsBERT}) drop to $\approx$0.21 macro-F\textsubscript{1} under full fine-tuning, while frozen-backbone MLP settings achieve $\approx$0.49. This suggests that, given the relatively small size of \textsc{PlayDial-FA}, full fine-tuning overfits and degrades pretrained representations.\\
\textbf{Remaining challenges.}
Despite having only three classes, theatrical sentiment remains difficult due to sarcasm, rhetorical exaggeration, and subtle polarity distinctions. The best model still reaches only 0.62 macro-F\textsubscript{1}, leaving substantial room for improvement.

\enlargethispage{2\baselineskip}
\section{Conclusion and Future Work}
\label{sec:conclusion}

\textsc{TalkFa} introduces the first unified benchmark for Farsi dialogue generation and understanding, combining \textsc{Wiki-FaDial}, \textsc{DailyDialog-FA}, and \textsc{PlayDial-FA}. Using LLM-assisted data creation with native-speaker verification, we establish reproducible baselines for generation, dialogue acts, emotions, and sentiment classification. LoRA substantially improves generation (+13 ROUGE-L $F_1$, $\sim$39\% BLEU-4) and consistently outperforms zero-shot settings. \textsc{FaBERT} achieves the best dialogue-act score (0.75 macro-F$_1$), \textsc{LoRA-Mistral-7B} the best emotion score (0.38), and LoRA-\textsc{Mistral-24B} the best sentiment result (0.62). Moreover, 25--50\% of the training data recovers over 90\% of final generation gains.
Human evaluation reveals that even the best fine-tuned model (\textsc{Mistral-Small-24B} +LoRA) achieves only 2.74/5 despite very high automatic scores (BERTCos = 0.97; BERTScore = 0.93), highlighting the limitations of reference-based metrics for open-ended dialogue generation. Frontier LLMs achieve competitive but non-saturating zero-shot performance, while GPT-4.1 as a judge shows moderate agreement with human ratings ($\rho = 0.43$, $\pm$1 = 79.7\%) but remains systematically more lenient.
We release all datasets, annotation guidelines, code, LoRA adapters, checkpoints, and baselines. Future work includes richer pragmatic annotations, retrieval-grounded dialogue, speech and code-switching extensions, adaptation to Dari/Tajik, and robustness and safety evaluation suites with a public leaderboard.

\section{Limitations}

While \textsc{TalkFa} advances evaluation for Farsi dialogue generation and understanding, several limitations remain.

\begin{itemize}

  \item \textbf{LLM-Assisted Data Construction and Potential Contamination.}
TalkFa was constructed with LLM assistance, using GPT-4o for dialogue generation, translation, segmentation, or reformulation depending on the dataset, followed by multi-stage human verification and post-editing by native speakers. Although this process substantially improves fluency, factual consistency, and cultural appropriateness, subtle stylistic regularities or lexical preferences inherited from the underlying model may still persist, potentially underrepresenting the diversity of fully human-authored conversations. In addition, some source materials, such as Wikipedia passages and public-domain theatrical scripts, may overlap with the pretraining corpora of contemporary LLMs. Because such overlap cannot be reliably verified, we do not claim that TalkFa is contamination-free. Future work could further investigate these effects by comparing TalkFa with newly available human-authored Farsi dialogue resources and by developing more rigorous contamination detection methods.
    \item \textbf{Structural Constraints in \textsc{Wiki-FaDial}:} 
    \textsc{Wiki-FaDial} uses a fixed six-turn structure with short utterances (1--20 tokens per turn). While this improves consistency and controllability, it underrepresents longer, multi-topic, overlapping, or multi-party conversations common in real-world dialogue. Models trained on the corpus may therefore become biased toward concise and orderly interactions.

    \item \textbf{Cross-Lingual Label Transfer in \textsc{DailyDialog-FA}:} 
    Dialogue-act and emotion labels are inherited from the English \textsc{DailyDialog} taxonomy. However, emotional expression and pragmatic intent in Farsi do not always align cleanly with English conversational categories. Phenomena such as \emph{ta'arof}, indirectness, politeness mitigation, and culturally dependent emotional expression may not be fully captured by the transferred label inventory.

    \item \textbf{Synthetic Reformulation in \textsc{PlayDial-FA}:} 
    Although \textsc{PlayDial-FA} originates from real theatrical scripts, dialogues are segmented and reformulated using \textsc{GPT-4o} to avoid verbatim reproduction and standardize formatting. This process may alter stylistic nuances, rhetorical structure, or author-specific writing patterns present in the original plays.

    \item \textbf{Limited Register and Dialect Coverage:} 
    Although \textsc{TalkFa} spans encyclopedic, everyday, and theatrical dialogue, it does not cover many important real-world settings such as social media interactions, spoken disfluencies, code-switching, online slang, or multi-party chats. The benchmark also focuses primarily on standard Iranian Farsi and does not systematically include regional or closely related varieties such as Dari or Tajik.

    \item \textbf{Evaluation Limitations:} 
    Despite incorporating human evaluation and LLM-as-a-judge analysis, dialogue evaluation remains inherently subjective. Human ratings may vary across annotators and cultural backgrounds, while LLM judges can exhibit systematic biases such as verbosity preference or leniency. In addition, automatic metrics (e.g., BLEU, ROUGE, BERTScore) correlate imperfectly with human judgments for open-ended conversational generation.

    \item \textbf{Benchmark Scale and Model Coverage:} 
    While \textsc{TalkFa} is substantially larger than prior Farsi dialogue resources, it remains smaller than major English conversational benchmarks. Similarly, our experiments cover representative open-source LLMs and encoders but do not exhaustively evaluate all architectures, prompting strategies, retrieval-based systems, or reasoning-oriented models.
    \item \textbf{Independent Validation Scope:}
Our external validation relied on a single independent annotator evaluating stratified samples of each task. While the resulting agreement was consistently high, future work could extend this evaluation by involving multiple independent annotators and larger validation sets to further assess annotation reproducibility.

\end{itemize}

\bibliography{custom}

@inproceedings{li2017dailydialog,
  title={Dailydialog: A manually labelled multi-turn dialogue dataset},
  author={Li, Yanran and Su, Hui and Shen, Xiaoyu and Li, Wenjie and Cao, Ziqiang and Niu, Shuzi},
  booktitle={Proceedings of the Eighth International Joint Conference on Natural Language Processing (Volume 1: Long Papers)},
  pages={986--995},
  year={2017}
}

@article{hu2022lora,
  title={Lora: Low-rank adaptation of large language models.},
  author={Hu, Edward J and Shen, Yelong and Wallis, Phillip and Allen-Zhu, Zeyuan and Li, Yuanzhi and Wang, Shean and Wang, Liang and Chen, Weizhu and others},
  journal={Iclr},
  volume={1},
  number={2},
  pages={3},
  year={2022}
}

@inproceedings{zhang2018personalizing,
  title={Personalizing dialogue agents: I have a dog, do you have pets too?},
  author={Zhang, Saizheng and Dinan, Emily and Urbanek, Jack and Szlam, Arthur and Kiela, Douwe and Weston, Jason},
  booktitle={Proceedings of the 56th Annual Meeting of the Association for Computational Linguistics (Volume 1: Long Papers)},
  pages={2204--2213},
  year={2018}
}

@inproceedings{xue2021mt5,
  title={mT5: A massively multilingual pre-trained text-to-text transformer},
  author={Xue, Linting and Constant, Noah and Roberts, Adam and Kale, Mihir and Al-Rfou, Rami and Siddhant, Aditya and Barua, Aditya and Raffel, Colin},
  booktitle={Proceedings of the 2021 conference of the North American chapter of the association for computational linguistics: Human language technologies},
  pages={483--498},
  year={2021}
}

@inproceedings{conneau2020unsupervised,
  title={Unsupervised cross-lingual representation learning at scale},
  author={Conneau, Alexis and Khandelwal, Kartikay and Goyal, Naman and Chaudhary, Vishrav and Wenzek, Guillaume and Guzm{\'a}n, Francisco and Grave, Edouard and Ott, Myle and Zettlemoyer, Luke and Stoyanov, Veselin},
  booktitle={Proceedings of the 58th annual meeting of the association for computational linguistics},
  pages={8440--8451},
  year={2020}
}

@misc{meta_llama3.2_1B_instruct,
  title = {meta-llama/Llama-3.2-1B-Instruct},
  author = {{Meta AI}},
  year = {2024},
  note = {HuggingFace model card},
  url = {https://huggingface.co/meta-llama/Llama-3.2-1B-Instruct}
}

@misc{meta_llama3.2_3B_instruct,
  title = {meta-llama/Llama-3.2-3B-Instruct},
  author = {{Meta AI}},
  year = {2024},
  note = {HuggingFace model card},
  url = {https://huggingface.co/meta-llama/Llama-3.2-3B-Instruct}
}

@misc{meta_llama3.1_8B_instruct,
  title = {meta-llama/Llama-3.1-8B-Instruct},
  author = {{Meta AI}},
  year = {2024},
  note = {HuggingFace model card},
  url = {https://huggingface.co/meta-llama/Llama-3.1-8B-Instruct}
}

@misc{mistral7b_instruct_v0.3,
  title = {mistralai/Mistral-7B-Instruct-v0.3},
  author = {{Mistral AI}},
  year = {2024},
  note = {HuggingFace model card},
  url = {https://huggingface.co/mistralai/Mistral-7B-Instruct-v0.3}
}

@misc{mistral_nemo_instruct_2407,
  title = {mistralai/Mistral-Nemo-Instruct-2407},
  author = {{Mistral AI}},
  year = {2024},
  note = {HuggingFace model card, developed with NVIDIA},
  url = {https://huggingface.co/mistralai/Mistral-Nemo-Instruct-2407}
}

@misc{mistral_small_24B_instruct_2501,
  title = {mistralai/Mistral-Small-24B-Instruct-2501},
  author = {{Mistral AI}},
  year = {2025},
  note = {HuggingFace model card},
  url = {https://huggingface.co/mistralai/Mistral-Small-24B-Instruct-2501}
}

@misc{intfloat_multilingual_e5_small,
  title = {intfloat/multilingual-e5-small},
  author = {{Microsoft}},
  year = {2024},
  note = {HuggingFace model card},
  url = {https://huggingface.co/intfloat/multilingual-e5-small}
}

@misc{intfloat_multilingual_e5_base,
  title = {intfloat/multilingual-e5-base},
  author = {{Microsoft}},
  year = {2024},
  note = {HuggingFace model card. Embedding model: 12 layers, embedding size 768},
  url = {https://huggingface.co/intfloat/multilingual-e5-base}
}

@misc{intfloat_multilingual_e5_large,
  title = {intfloat/multilingual-e5-large},
  author = {{Microsoft}},
  year = {2024},
  note = {HuggingFace model card. Embedding model: 24 layers, embedding size 1024; supports about 100 languages},
  url = {https://huggingface.co/intfloat/multilingual-e5-large}
}

@misc{hooshvarelab_bert_base_parsbert_uncased,
  title = {HooshvareLab/bert-base-parsbert-uncased},
  author = {{HooshvareLab}},
  year = {2020},
  note = {HuggingFace model card. Monolingual Persian BERT model (ParsBERT)},
  url = {https://huggingface.co/HooshvareLab/bert-base-parsbert-uncased}
}

@misc{sbunlp_fabert,
  title = {sbunlp/fabert},
  author = {{Shahid Beheshti University NLP Group}},
  year = {2024},
  note = {HuggingFace model card. Persian BERT-base model (FaBERT) trained on HmBlogs corpus},
  url = {https://huggingface.co/sbunlp/fabert}
}

@misc{google_bert_base_multilingual_cased,
  title = {google-bert/bert-base-multilingual-cased},
  author = {{Google}},
  year = {2024},
  note = {HuggingFace model card. Multilingual BERT trained on 104 Wikipedia languages},
  url = {https://huggingface.co/google-bert/bert-base-multilingual-cased}
}

@misc{facebookai_roberta_base,
  title = {FacebookAI/roberta-base},
  author = {{Meta AI}},
  year = {2024},
  note = {HuggingFace model card. Pre-trained English RoBERTa-base model},
  url = {https://huggingface.co/FacebookAI/roberta-base}
}

@misc{facebookai_roberta_large,
  title = {FacebookAI/roberta-large},
  author = {{Meta AI}},
  year = {2024},
  note = {HuggingFace model card. RoBERTa-large model introduced in the RoBERTa paper},
  url = {https://huggingface.co/FacebookAI/roberta-large}
}

@misc{facebookai_xlm-roberta_base,
  title = {FacebookAI/xlm-roberta-base},
  author = {{Meta AI}},
  year = {2019},
  note = {HuggingFace model card. Multilingual RoBERTa pretrained on CommonCrawl (100 languages)},
  url = {https://huggingface.co/FacebookAI/xlm-roberta-base}
}

@misc{facebookai_xlm-roberta_large,
  title = {FacebookAI/xlm-roberta-large},
  author = {{Meta AI}},
  year = {2024},
  note = {HuggingFace model card. Large multilingual XLM-RoBERTa model},
  url = {https://huggingface.co/FacebookAI/xlm-roberta-large}
}

@misc{hooshvarelab_bert_fa_base_uncased,
  title = {HooshvareLab/bert-fa-base-uncased},
  author = {{HooshvareLab}},
  year = {2020},
  note = {HuggingFace model card. Persian language model ParsBERT v2.0},
  url = {https://huggingface.co/HooshvareLab/bert-fa-base-uncased}
}

@misc{wikipedia_fa,
  title        = {Featured Articles in Persian Wikipedia},
  author       = {{Wikipedia contributors}},
  year         = {2025},
  howpublished = {\url{https://fa.wikipedia.org}},
}

@misc{100_essential_articles,
  title  = {100 Essential Articles (Persian Wikipedia)},
  author = {{Wikipedia contributors}},
  year   = {2025},
  url    = {https://fa.wikipedia.org/wiki/Wikipedia:100_essential_articles}
}

@misc{essential_articles_every_wiki_should_have,
  title        = {Essential Articles Every Wikipedia Should Have},
  author       = {{Wikipedia contributors}},
  year         = {2009},
  howpublished = {\url{https://fa.wikipedia.org}}
}

@misc{most_viewed,
  title  = {Most Viewed Articles in Persian Wikipedia},
  author = {{Wikipedia contributors}},
  year   = {2014},
  url    = {https://fa.wikipedia.org}
}

@misc{takbook_com,
  title = {TakBook: Free Online Book Download Platform},
  author = {{TakBook}},
  year = {2025},
  note = {Website},
  url = {https://www.takbook.com/}
}

@misc{pdf2go_pdf_to_text,
  title = {PDF2Go: Convert PDF to Text Online},
  author = {{PDF2Go}},
  year = {2025},
  note = {Online PDF conversion tool},
  url = {https://www.pdf2go.com/pdf-to-text}
}

@inproceedings{budzianowski2018multiwoz,
  title={Multiwoz-a large-scale multi-domain wizard-of-oz dataset for task-oriented dialogue modelling},
  author={Budzianowski, Pawe{\l} and Wen, Tsung-Hsien and Tseng, Bo-Hsiang and Casanueva, I{\~n}igo and Ultes, Stefan and Ramadan, Osman and Gasic, Milica},
  booktitle={Proceedings of the 2018 conference on empirical methods in natural language processing},
  pages={5016--5026},
  year={2018}
}

@article{gopalakrishnan2023topical,
  title={Topical-chat: Towards knowledge-grounded open-domain conversations},
  author={Gopalakrishnan, Karthik and Hedayatnia, Behnam and Chen, Qinlang and Gottardi, Anna and Kwatra, Sanjeev and Venkatesh, Anu and Gabriel, Raefer and Hakkani-Tur, Dilek},
  journal={arXiv preprint arXiv:2308.11995},
  year={2023}
}

@article{rashkin2018towards,
  title={Towards empathetic open-domain conversation models: A new benchmark and dataset},
  author={Rashkin, Hannah and Smith, Eric Michael and Li, Margaret and Boureau, Y-Lan},
  journal={arXiv preprint arXiv:1811.00207},
  year={2018}
}

@article{li2020mtop,
  title={MTOP: A comprehensive multilingual task-oriented semantic parsing benchmark},
  author={Li, Haoran and Arora, Abhinav and Chen, Shuohui and Gupta, Anchit and Gupta, Sonal and Mehdad, Yashar},
  journal={arXiv preprint arXiv:2008.09335},
  year={2020}
}

@article{khashabi2021parsinlu,
  title={Parsinlu: a suite of language understanding challenges for persian},
  author={Khashabi, Daniel and Cohan, Arman and Shakeri, Siamak and Hosseini, Pedram and Pezeshkpour, Pouya and Alikhani, Malihe and Aminnaseri, Moin and Bitaab, Marzieh and Brahman, Faeze and Ghazarian, Sarik and others},
  journal={Transactions of the Association for Computational Linguistics},
  volume={9},
  pages={1147--1162},
  year={2021}
}

@inproceedings{sadallah2025commonsense,
  title={Commonsense reasoning in arab culture},
  author={Sadallah, Abdelrahman and Tonga, Junior Cedric and Almubarak, Khalid and Almheiri, Saeed and Atif, Farah and Qwaider, Chatrine and Kadaoui, Karima and Shatnawi, Sara and Alesh, Yaser and Koto, Fajri},
  booktitle={Proceedings of the 63rd Annual Meeting of the Association for Computational Linguistics (Volume 1: Long Papers)},
  pages={7695--7710},
  year={2025}
}

@inproceedings{purwarianti2025nusadialogue,
  title={NusaDialogue: Dialogue summarization and generation for underrepresented and extremely low-resource languages},
  author={Purwarianti, Ayu and Adhista, Dea and Baptiso, Agung and Mahfuzh, Miftahul and Sabila, Yusrina and Adila, Aulia and Cahyawijaya, Samuel and Aji, Alham Fikri},
  booktitle={Proceedings of the Second Workshop in South East Asian Language Processing},
  pages={82--100},
  year={2025}
}

@article{shamsfard2025farseval,
  title={FarsEval-PKBETS: A new diverse benchmark for evaluating Persian large language models},
  author={Shamsfard, Mehrnoush and Saaberi, Zahra and Hashemi, Seyed Mohammad Hossein and Vatankhah, Zahra and Ramezani, Motahareh and Pourazin, Niki and Zare, Tara and Azimi, Maryam and Chitsaz, Sarina and Khoraminejad, Sama and others},
  journal={arXiv preprint arXiv:2504.14690},
  year={2025}
}

@inproceedings{suresh2025diasynth,
  title={Diasynth: Synthetic dialogue generation framework for low resource dialogue applications},
  author={Suresh, Sathya Krishnan and Mengjun, Wu and Pranav, Tushar and Chng, Eng Siong},
  booktitle={Findings of the Association for Computational Linguistics: NAACL 2025},
  pages={673--690},
  year={2025}
}

@article{dinan2018wizard,
  title={Wizard of wikipedia: Knowledge-powered conversational agents},
  author={Dinan, Emily and Roller, Stephen and Shuster, Kurt and Fan, Angela and Auli, Michael and Weston, Jason},
  journal={arXiv preprint arXiv:1811.01241},
  year={2018}
}

@article{dziri2022faithdial,
  title={Faithdial: A faithful benchmark for information-seeking dialogue},
  author={Dziri, Nouha and Kamalloo, Ehsan and Milton, Sivan and Zaiane, Osmar and Yu, Mo and Ponti, Edoardo M and Reddy, Siva},
  journal={Transactions of the Association for Computational Linguistics},
  volume={10},
  pages={1473--1490},
  year={2022},
  publisher={MIT Press One Broadway, 12th Floor, Cambridge, Massachusetts 02142, USA~…}
}

@article{dziri2022evaluating,
  title={Evaluating attribution in dialogue systems: The BEGIN benchmark},
  author={Dziri, Nouha and Rashkin, Hannah and Linzen, Tal and Reitter, David},
  journal={Transactions of the Association for Computational Linguistics},
  volume={10},
  pages={1066--1083},
  year={2022},
  publisher={MIT Press One Broadway, 12th Floor, Cambridge, Massachusetts 02142, USA~…}
}

@inproceedings{zhou2020kdconv,
  title={KdConv: A Chinese multi-domain dialogue dataset towards multi-turn knowledge-driven conversation},
  author={Zhou, Hao and Zheng, Chujie and Huang, Kaili and Huang, Minlie and Zhu, Xiaoyan},
  booktitle={Proceedings of the 58th Annual Meeting of the Association for Computational Linguistics},
  pages={7098--7108},
  year={2020}
}

@inproceedings{wu2019proactive,
  title={Proactive human-machine conversation with explicit conversation goal},
  author={Wu, Wenquan and Guo, Zhen and Zhou, Xiangyang and Wu, Hua and Zhang, Xiyuan and Lian, Rongzhong and Wang, Haifeng},
  booktitle={Proceedings of the 57th Annual Meeting of the Association for Computational Linguistics},
  pages={3794--3804},
  year={2019}
}

@inproceedings{zhang2023xdial,
  title={xDial-eval: A multilingual open-domain dialogue evaluation benchmark},
  author={Zhang, Chen and D’Haro, Luis and Tang, Chengguang and Shi, Ke and Tang, Guohua and Li, Haizhou},
  booktitle={Findings of the Association for Computational Linguistics: EMNLP 2023},
  pages={5579--5601},
  year={2023}
}

@inproceedings{ahuja2023mega,
  title={Mega: Multilingual evaluation of generative ai},
  author={Ahuja, Kabir and Diddee, Harshita and Hada, Rishav and Ochieng, Millicent and Ramesh, Krithika and Jain, Prachi and Nambi, Akshay and Ganu, Tanuja and Segal, Sameer and Ahmed, Mohamed and others},
  booktitle={Proceedings of the 2023 Conference on Empirical Methods in Natural Language Processing},
  pages={4232--4267},
  year={2023}
}

@inproceedings{liu2016not,
  title={How not to evaluate your dialogue system: An empirical study of unsupervised evaluation metrics for dialogue response generation},
  author={Liu, Chia-Wei and Lowe, Ryan and Serban, Iulian Vlad and Noseworthy, Mike and Charlin, Laurent and Pineau, Joelle},
  booktitle={Proceedings of the 2016 conference on empirical methods in natural language processing},
  pages={2122--2132},
  year={2016}
}

@article{zhang2019bertscore,
  title={Bertscore: Evaluating text generation with bert},
  author={Zhang, Tianyi and Kishore, Varsha and Wu, Felix and Weinberger, Kilian Q and Artzi, Yoav},
  journal={arXiv preprint arXiv:1904.09675},
  year={2019}
}

@inproceedings{papineni2002bleu,
  title={Bleu: a method for automatic evaluation of machine translation},
  author={Papineni, Kishore and Roukos, Salim and Ward, Todd and Zhu, Wei-Jing},
  booktitle={Proceedings of the 40th annual meeting of the Association for Computational Linguistics},
  pages={311--318},
  year={2002}
}

@inproceedings{lin2004rouge,
  title={Rouge: A package for automatic evaluation of summaries},
  author={Lin, Chin-Yew},
  booktitle={Text summarization branches out},
  pages={74--81},
  year={2004}
}

@inproceedings{mehri2020usr,
  title={USR: An unsupervised and reference free evaluation metric for dialog generation},
  author={Mehri, Shikib and Eskenazi, Maxine},
  booktitle={Proceedings of the 58th Annual Meeting of the Association for Computational Linguistics},
  pages={681--707},
  year={2020}
}

@article{zheng2023judging,
  title={Judging llm-as-a-judge with mt-bench and chatbot arena},
  author={Zheng, Lianmin and Chiang, Wei-Lin and Sheng, Ying and Zhuang, Siyuan and Wu, Zhanghao and Zhuang, Yonghao and Lin, Zi and Li, Zhuohan and Li, Dacheng and Xing, Eric and others},
  journal={Advances in neural information processing systems},
  volume={36},
  pages={46595--46623},
  year={2023}
}

@inproceedings{lee2023making,
  title={Making large language models better data creators},
  author={Lee, Dong-Ho and Pujara, Jay and Sewak, Mohit and White, Ryen and Jauhar, Sujay},
  booktitle={Proceedings of the 2023 Conference on Empirical Methods in Natural Language Processing},
  pages={15349--15360},
  year={2023}
}

@inproceedings{zhou2018dataset,
  title={A dataset for document grounded conversations},
  author={Zhou, Kangyan and Prabhumoye, Shrimai and Black, Alan W},
  booktitle={Proceedings of the 2018 conference on empirical methods in natural language processing},
  pages={708--713},
  year={2018}
}

@article{landis1977measurement,
  title={The measurement of observer agreement for categorical data},
  author={Landis, J Richard and Koch, Gary G},
  journal={biometrics},
  pages={159--174},
  year={1977},
  publisher={JSTOR}
}

@inproceedings{mehri2020unsupervised,
  title={Unsupervised evaluation of interactive dialog with DialoGPT},
  author={Mehri, Shikib and Eskenazi, Maxine},
  booktitle={Proceedings of the 21th Annual Meeting of the Special Interest Group on Discourse and Dialogue},
  pages={225--235},
  year={2020}
}

@misc{deepseek_v4_flash,
  author       = {{DeepSeek-AI}},
  title        = {DeepSeek-V4-Flash},
  year         = {2026},
  howpublished = {Hugging Face},
  url          = {https://huggingface.co/deepseek-ai/DeepSeek-V4-Flash},
  note         = {Model card}
}

@misc{openai2025gpt41,
  author       = {{OpenAI}},
  title        = {Introducing GPT-4.1 in the API},
  year         = {2025},
  month        = apr,
  howpublished = {OpenAI},
  url          = {https://openai.com/index/gpt-4-1/},
 
}

@misc{openai2024gpt4o,
  author       = {{OpenAI}},
  title        = {Hello GPT-4o},
  year         = {2024},
  month        = may,
  howpublished = {OpenAI},
  url          = {https://openai.com/index/hello-gpt-4o/},
  
}

\clearpage          %

\appendix
\twocolumn
\section{Full Dialogues of \textsc{Wiki-FaDial}, \textsc{DailyDialog-FA}, and \textsc{PlayDial-FA}}

\label{app:examples}
Here, we present the complete dialogues from the three sources illustrated in Figure~\ref{fig:Figure1}: \textsc{Wiki-FaDial} (natural dialogues generated from Farsi Wikipedia paragraphs), \textsc{DailyDialog-FA} (daily English conversations translated into culturally natural Farsi), and \textsc{PlayDial-FA} (expressive multi-turn dialogues extracted from Farsi plays).

\includepdf[pages=-]{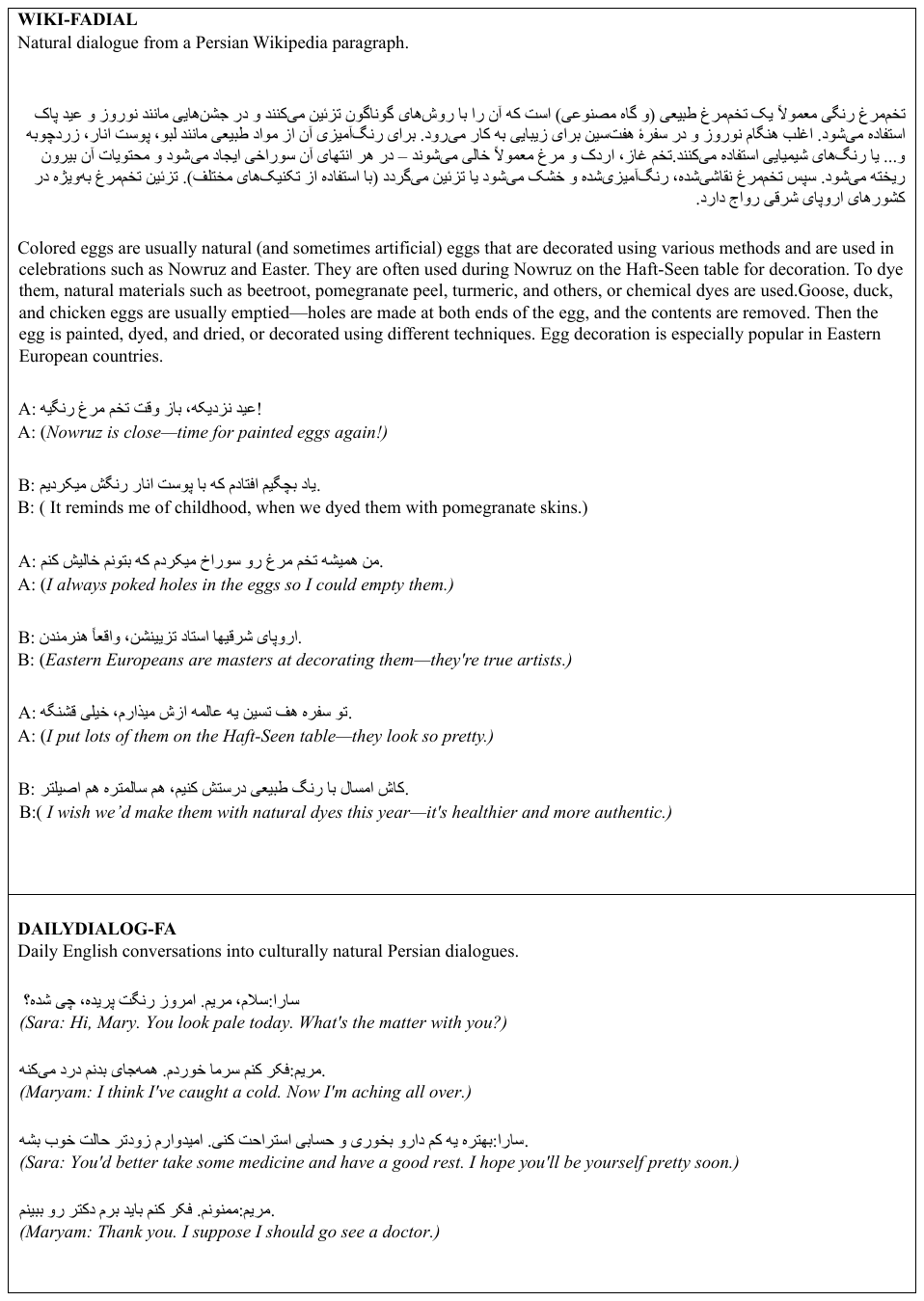}

\twocolumn
\section{Prompt Used for \textsc{Wiki-FaDial} Generation}
\label{app:promptwiki}

\paragraph{System prompt:}
This GPT model takes Farsi texts and transforms each one into a casual, informal dialogue between two people. The generated dialogue should:
\begin{itemize}
    \item be natural, relatable, and culturally appropriate in Farsi,
    \item include elements such as humor, honest reflection, light-hearted frustration, empathetic advice, and small talk,
    \item reflect personal or work-related topics and daily routines when appropriate,
    \item use natural idioms or slang and maintain an informal tone,
    \item contain exactly 6 turns, each between 1 and 20 words,
    \item include a \texttt{"reference"} field in each turn quoting or summarizing the relevant part of the source text for transparency.
\end{itemize}

\paragraph{Input format (JSON):}
\begin{verbatim}
{
  "texts": [
    "First Farsi text...",
    "Second Farsi text...",
    "Third Farsi text...",
    "Fourth Farsi text...",
    "Fifth Farsi text..."
  ]
}
\end{verbatim}

\paragraph{Output format:}
The model produces five independent JSON objects (one per input text). Each object contains only the generated dialogue and does not include the original input text:

\begin{verbatim}
{
  "dialogue": [
    {
      "speaker": "A",
      "text": "...",
      "reference": "..."
    },
    {
      "speaker": "B",
      "text": "...",
      "reference": "..."
    }
  ]
}
\end{verbatim}

\twocolumn

\twocolumn
\section{Human Post-editing Protocol for \textsc{Wiki-FaDial} }
\label{app:guidelines_wiki}

Below is the streamlined protocol followed by native-speaker annotators when refining \textsc{Wiki-FaDial} dialogues.  
Two annotators work independently; a senior linguist resolves any conflicts.

\begin{quote}

\textbf{Guideline:} When editing or regenerating these dialogues, make sure each has exactly 6 short turns (each 1–20 words) in casual, friendly, natural Farsi, sounding like two people in a real-life chat. The language must be informal, sprinkled with humor, light frustration, empathy, or small talk, using idioms and spoken expressions that feel truly Iranian.

Each turn must include a reference that ties it clearly to the original text (a short quote or idea summary) to show transparency. Ensure the conversation is not a dry summary but a playful, human back-and-forth — like two friends reacting, joking, giving quick advice, or sharing small relatable complaints.

After generating, carefully check the dialogue for all these points: if it fails (too stiff, formal, robotic, or missing references), regenerate it with a prompt asking for playful, informal style, or manually fix grammar, shorten long lines, replace unnatural words, or add emotional touches until it feels authentic.

Throughout, keep the JSON clean (no trailing commas, correct brackets), so the final output is short, vivid, idiomatic, well-referenced, and technically valid.
\end{quote}

\subsection*{Stage 1 – Coherence \& Flow}
\begin{itemize}
  \item Ensure each question follows logically from the previous answer.
  \item Delete turns that introduce off-topic content or repeat earlier facts.
\end{itemize}

\subsection*{Stage 2 – Factuality \& Reference}
\begin{itemize}
  \item Check every statement against the source paragraph; replace or remove hallucinations.
  \item Verify names, dates, and numbers; correct diacritics.
\end{itemize}

\subsection*{Stage 3 – Style \& Register}
\begin{itemize}
  \item Keep each turn 1--20 tokens long; split or condense otherwise.
  \item Use informal Farsi: swap formal forms for idioms/slang; add discourse markers sparingly.
  \item Optional: insert light humour, empathy, or advice—without altering facts.
\end{itemize}

\subsection*{Stage 4 – Compliance \& Safety}
\begin{itemize}
  \item check Hate checks; anonymise or delete flagged text.
  \item Remove any content violating Wikimedia policy (explicit, medical, self-harm).
\end{itemize}

\section*{Reproducibility Details}
\label{sec:reproducibility}

\paragraph{Hardware.}  
Experiments are conducted on \textbf{3$\times$ NVIDIA RTX A6000} GPUs (48\,GB each) with mixed-precision (bf16), FlashAttention, and gradient checkpointing.

\paragraph{Software environment.}
Python 3.11, PyTorch 2.3.0, HuggingFace Transformers $\ge$ 0.22, and BitsAndBytes 0.45 for 4-bit QLoRA.  All package versions are pinned in \texttt{environment.yml}.

\paragraph{Random seeds.}
We fix \texttt{seed = 100} for model initialisation, dataloader shuffling, and NumPy/PyTorch RNGs.  Results are averaged over three seeds for classification.

\paragraph{Hyper-parameters.}
\begin{itemize}
  \item \emph{LLM fine-tuning (LoRA):} per-device batch size = 4, gradient accumulation = 2 (effective batch size = 24 across 3 GPUs), lr =~\(1\!\times\!10^{-4}\) (AdamW, cosine decay), weight decay =~\(1\!\times\!10^{-4}\), warmup ratio = 0, max grad-norm = 1.0, epochs = 3, context = 512 tokens. LoRA configuration: rank = 128, $\alpha$ = 256, dropout = 0.1.
  \item \emph{Encoder classification (Frozen encoder + MLP head):} batch = 64, lr =~\(1\!\times\!10^{-5}\) (AdamW), epochs = 100, context = 256 tokens. 4-layer head: 512, 256, 128, 64 $\rightarrow$ \texttt{LogSoftmax}, NLLLoss.
  \item \emph{Encoder classification (Full fine-tuning):} batch = 32, lr =~\(2\!\times\!10^{-5}\) (AdamW), epochs = 3, context = 256 tokens.
\end{itemize}

\paragraph{Code \& data.}
All corpora, scripts, and checkpoints will be released.

\paragraph{Compute footprint.}
LLM fine-tuning consumes 16 GPU-hours in total; encoder runs finish in \(\sim\)2 hours per task.  This modest budget enables replication on most academic clusters.

\section{Illustrative Examples of Human Post-editing for \textsc{Wiki-FaDial}}
\label{app:editing_example}
Below are five examples of human post-editing for \textsc{Wiki-FaDial}. Each example includes the original Wikipedia passage, the dialogue generated by the model, and the final version after manual revision by human annotators.The following examples illustrate different types of interventions:

\begin{itemize}
    \item \textbf{Example 1:} The opening sentence contained a grammatical error that obscured the intended meaning, while the remainder of the dialogue was fluent and well-structured. As a result, only the first sentence was manually revised.

    \item \textbf{Example 2:} The model overused repetitive openers such as ``Have you heard...'' or ``Have you read...''. To increase conversational variety and avoid redundancy, annotators modified these introductions manually.

    \item \textbf{Example 3:} The generated sentences lacked syntactic coherence and logical progression, resulting in a disjointed and confusing dialogue. Consequently, annotators chose to regenerate the entire dialogue using the model.

    \item \textbf{Example 4:} The model produced an inaccurate and unrealistic analogy—e.g., comparing a storm to a funnel—that was not grounded in the original source content. This type of error, commonly seen in metaphorical or comparative contexts, was corrected by regenerating the dialogue, which yielded a coherent and accurate result on the second attempt.

    \item \textbf{Example 5:} No revisions were required. The model-generated dialogue was grammatically sound, culturally appropriate, and naturally reflected Farsi conversational style, even capturing a light sense of humor typical of everyday speech.
\end{itemize}

\includepdf[pages=-]{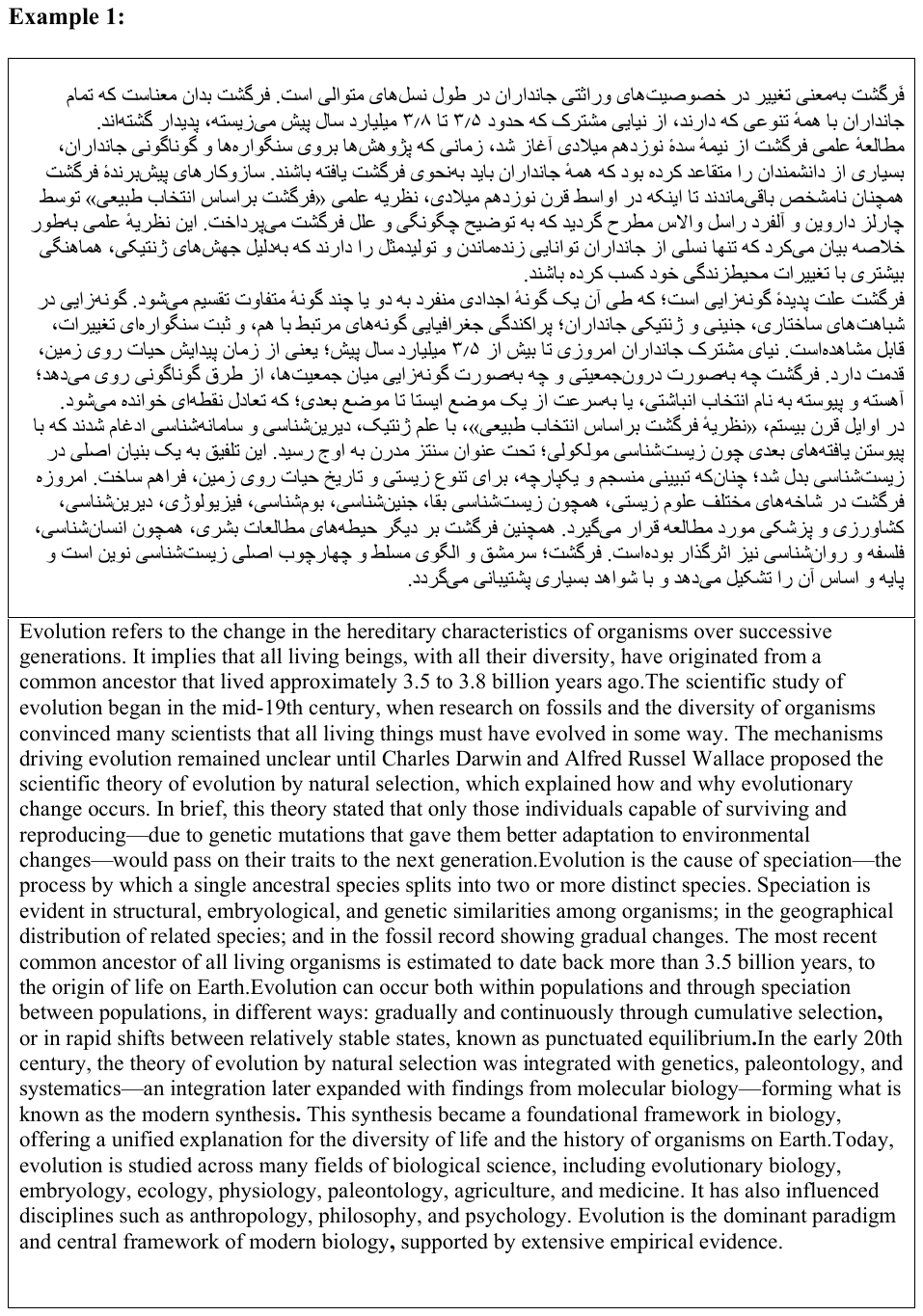}

\section{Model Checkpoints}
\label{app:model_ids}

\paragraph{Large-language models.}
\begin{enumerate}
  \item \texttt{meta-llama/Llama-3.2-1B-Instruct}
  \item \texttt{meta-llama/Llama-3.2-3B-Instruct}
  \item \texttt{meta-llama/Llama-3.1-8B-Instruct}
  \item \texttt{mistralai/Mistral-7B-Instruct-v0.3}
  \item \texttt{mistralai/Mistral-Nemo-Instruct-2407}
  \item \texttt{mistralai/Mistral-Small-24B-Instruct-2501}
\end{enumerate}

\paragraph{Transformer encoders.}
\begin{enumerate}
  \item \texttt{intfloat/multilingual-e5-large}
  \item \texttt{intfloat/multilingual-e5-base}
  \item \texttt{intfloat/multilingual-e5-small}
  \item \texttt{HooshvareLab/bert-base-parsbert-uncased}
  \item \texttt{sbunlp/fabert}
  \item \texttt{google-bert/bert-base-multilingual-cased}
  \item \texttt{FacebookAI/roberta-base}
  \item \texttt{FacebookAI/roberta-large}
  \item \texttt{FacebookAI/xlm-roberta-large}
  \item \texttt{FacebookAI/xlm-roberta-base}
\end{enumerate}

\section*{Translation Prompt}
\label{dialoug:prompt}

The following prompt was used to generate culturally authentic Farsi dialogues from English sources:

\begin{quote}
\small
Your task is to translate a given dialogue from English to Farsi, ensuring the following guidelines are followed:

1. Farsi Names: Use Farsi names in the translated dialogue while maintaining the gender distinctions present in the English version.

2. Natural Language: Craft the conversation in everyday spoken Farsi, utilizing common expressions and adhering to cultural norms of politeness.

3. Engaging Flow: Ensure the dialogue consists of at least six exchanges between two characters. Maintain a natural, logical, and engaging flow. Where relevant, include elements such as questions, disagreements, or problem-solving.

4. Cultural Authenticity: Reflect authentic Farsi cultural nuances in the conversation.

5. Dialogue Length: Keep the length of the dialogue the same as in the English version. Do not add any new conversations or exchanges.

6. Output Format: Provide the complete conversation in a valid JSON format. The dialogue should be entirely in Farsi, with no additional English commentary or context. Maintain logical coherence and consistency throughout.

\vspace{0.5em}
\textbf{Expected JSON Output Format:}
\begin{verbatim}
{
  "conversation": [
    {
      "speaker": "first person",
      "text": "{{response_1}}"
    },
    ...
  ]
}
\end{verbatim}

\textbf{English dialogue:} {dialog}
\end{quote}

\section{Human Post-editing Protocol for  \textsc{DailyDialog-FA}}
\label{app:guidelines_translate}
Below is the streamlined protocol followed by native-speaker annotators when refining \textit{DailyDialog-FA} dialogues. Two annotators work independently to post-edit the model-generated Farsi dialogues to ensure naturalness, cultural relevance, and structural fidelity to the original English source.

\subsection*{Editing Procedure}

Once the model generates the Farsi dialogue from English input, human annotators follow this step-by-step process:

\begin{enumerate}
    \item \textbf{Fluency and Naturalness} \\
    Ensure that each utterance reads like natural, everyday spoken Farsi. Sentences should flow conversationally, avoiding overly literal translations or stiff, unnatural phrasing. Aim for speech that reflects how native speakers actually talk.

    \item \textbf{Cultural Localization} \\
    Adapt all cultural elements to fit an Iranian context. This includes:
    \begin{itemize}
        \item Replacing foreign currencies (e.g., USD, Euro) with the appropriate Iranian currency.
        \item Substituting references to globally known locations (e.g., Niagara Falls) with locally recognizable or culturally familiar places.
        \item Adjusting units of measurement or public references (e.g., holidays, social norms) to culturally meaningful equivalents.
    \end{itemize}

    \item \textbf{Preserving Dialogue Structure} \\
    Keep the number of dialogue turns exactly the same as the original English version. Do not split, merge, or remove turns unless absolutely necessary for clarity or fluency. Preserve the logical progression and intention behind each exchange.

    \item \textbf{Speaker Identity and Naming} \\
    Assign culturally appropriate Farsi names that match the gender and style of the original English characters. Avoid names that may seem foreign or uncommon in the Iranian context. Gender agreement between speakers’ names and pronouns must be consistent throughout.

    \item \textbf{Grammatical and Semantic Corrections} \\
    Correct any errors in grammar, word choice, or sentence structure. Fix mistranslations that may alter or obscure the intended meaning. Pay special attention to:
    \begin{itemize}
        \item Verb tense consistency
        \item Proper use of prepositions and conjunctions
        \item Subject-verb agreement
        \item Pronoun resolution
    \end{itemize}

    \item \textbf{Removal of Artifacts} \\
    Eliminate any residual English words, incomplete phrases, formatting markers, or annotation artifacts (e.g., brackets, placeholder tokens). Ensure that the dialogue reads cleanly, without machine-generated traces.

    \item \textbf{Tone and Politeness} \\
    Maintain an appropriate register for informal dialogue. The tone should feel friendly, polite, and contextually appropriate for casual interaction. Avoid archaic, poetic, or overly formal language unless explicitly present in the original.

    \item \textbf{Formatting and Output} \\
    Ensure the final dialogue is submitted in valid JSON format, with clearly structured key-value pairs for each turn. No syntax errors should be present, and spacing or punctuation should follow standard conventions.
\end{enumerate}

\section{Illustrative Examples of Human Post-editing for  \textsc{DailyDialog-FA}}
\label{app:editing_example_translate}

Below are five representative examples from  \textsc{DailyDialog-FA}. Each example includes the original English dialogue, the raw Farsi translation generated by the model, and the final version after human post-editing. These examples illustrate the types of edits performed and the reasoning behind them.

\begin{itemize}
    \item \textbf{Example 1:} References to “dollars” were converted to “tomans” to reflect the local currency used in Iran.
    \item \textbf{Example 2:} A grading scale out of 100 was adapted to a 0–20 scale, which is standard in the Iranian education system.
    \item \textbf{Example 3:} Mention of China’s Huangguoshu Waterfall was replaced with Iran’s Laton Waterfall to create a culturally familiar context.
    \item \textbf{Example 4:} An American store name was substituted with a well-known Iranian brand.
    \item \textbf{Example 5:} The model autonomously replaced an American personal name with a natural-sounding Farsi name, requiring no human edits.
\end{itemize}

Each case illustrates the specific types of transformations necessary to produce high-quality, culturally grounded dialogues that are suitable for native Farsi speakers.

\includepdf[pages=-]{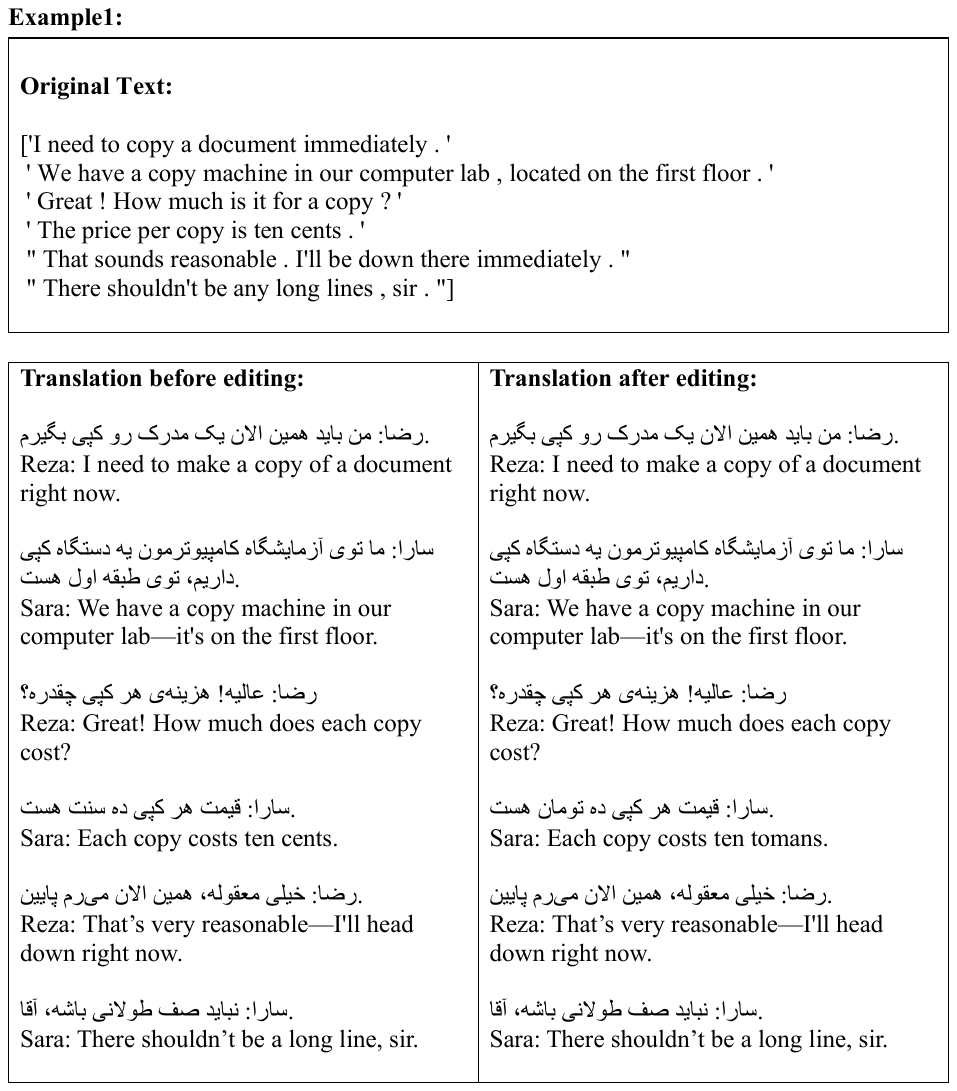}

\section{Segmentation prompt}
\label{app:segmentation-prompt}

\subsection*{Task Description}
Your duty is to receive the text and segment it strictly based on the rules below.  
The output must be in \textbf{Farsi} and presented in a \textbf{clear, organized table format}.

\subsection*{Segmentation Rules}
\begin{enumerate}
  \item \textbf{Scene Continuity:}
  \begin{itemize}
    \item Strictly follow scene continuity.
    \item Start a new segment only when there is an \textbf{explicit scene change}, such as:
    \begin{itemize}
      \item Change in \textbf{location}
      \item Change in \textbf{time}
      \item A significant shift in the \textbf{conversation context} (not a minor one)
    \end{itemize}
  \end{itemize}

  \item \textbf{Dialogue Preservation:}
  \begin{itemize}
    \item Preserve the original dialogues \textbf{exactly as they appear in the text}.
    \item Do \textbf{not} omit, alter, summarize, or paraphrase any part.
    \item Keep interruptions, incomplete sentences, repetitions, and pauses.
    \item Clearly indicate the \textbf{speaker} for each line.
    \item Double-check for completeness to ensure no dialogue lines are missing after segmentation.
  \end{itemize}

  \item \textbf{Output Format:}
  \begin{itemize}
    \item Present the output in a table with the following columns:
    \begin{enumerate}
      \item \textbf{Title:} A short, descriptive title summarizing the essence of the scene.
      \item \textbf{Characters:} List all characters who have dialogues or are mentioned in the segment.
      \item \textbf{Dialogues:} Present all dialogues exactly as in the original script, clearly marking each speaker.
      \item \textbf{Overall Sentiment:} Choose from \textit{Positive}, \textit{Negative}, or \textit{Neutral}.
      \item \textbf{Reference:} Always write the name of the theatre play.
    \end{enumerate}
  \end{itemize}
\end{enumerate}

\subsection*{Additional Instructions}
\begin{itemize}
  \item Ensure each scene is \textbf{complete} before moving to the next one.
  \item Do \textbf{not} omit any part of the text.
  \item Double-check for any missing dialogues after segmentation.
  \item Maintain clarity and organization for easy analysis.
\end{itemize}

\section{Annotation and Editing Protocol for Play-Based Dialogues}
\label{app:play_guidelines}

To ensure consistency and quality during corpus construction, annotators followed a structured protocol while cleaning, extracting, reformulating, and validating dialogue segments from the Farsi play scripts.

\begin{itemize}

\item \textbf{Remove non-dialogue content.}
Delete stage directions, narrator descriptions, scene headers, page numbers, and formatting artifacts introduced by OCR. Only spoken dialogue lines should remain.

\item \textbf{Preserve original dialogue content.}
The objective of the extraction stage is to retain the spoken dialogue exactly as it appears in the original plays. Annotators must not paraphrase, rewrite, or introduce new content at this stage.

\item \textbf{Correct OCR artifacts.}
Fix obvious OCR errors such as broken characters, incorrect punctuation, or merged tokens while preserving the intended wording of the original script.

\item \textbf{Maintain speaker attribution.}
Ensure that each dialogue line remains associated with the correct character in the original script. If OCR errors obscure the speaker name, the annotator should verify it by consulting the original PDF.

\item \textbf{Preserve conversational order.}
Dialogue lines must remain in the exact sequence in which they appear in the original play. Reordering dialogue turns is not permitted.

\item \textbf{Remove stage interruptions within dialogue.}
If stage directions appear inside dialogue lines (e.g., gestures, pauses, or descriptions of actions), remove them while keeping the surrounding spoken text intact.

\item \textbf{Verify extraction against the original script.}
After cleaning the OCR output, annotators compare the extracted dialogue with the original play to ensure that no dialogue lines were accidentally omitted or altered.

\item \textbf{Check scene boundaries for segmentation.}
When scripts are segmented into dialogue scenes, ensure that a new dialogue begins only when the scene clearly changes in the original play. Segmentation should preserve the natural conversational flow within each scene.

\item \textbf{Verify reformulated dialogues for copyright compliance.}
After segmentation, dialogue segments are reformulated using \textsc{GPT-4o} to avoid verbatim reproduction of copyrighted play scripts. Annotators compare each reformulated dialogue with the original scene to ensure that the conversational intent, speaker interactions, and thematic content are preserved while avoiding direct copying of the source text.

\item \textbf{Preserve conversational structure.}
During reformulation verification, annotators ensure that the number of turns and the overall conversational flow remain consistent with the original scene.

\item \textbf{Prevent verbatim overlap.}
Annotators ensure that reformulated dialogues do not contain long verbatim spans from the original plays, thereby preventing redistribution of copyrighted text.

\item \textbf{Flag ambiguous or corrupted cases.}
If OCR errors, missing dialogue, or unclear speaker attribution are encountered, annotators flag the issue for discussion and resolve it during the adjudication stage.

\item \textbf{Validate sentiment labels.}
For sentiment annotation, annotators review the dialogue within the context of the original scene to determine whether the overall sentiment is \emph{positive}, \emph{neutral}, or \emph{negative}. When disagreements occur, annotators discuss the case and finalize the label through consensus.

\end{itemize}

\section{Example of Segmentation, Reformulation, and Human Verification}
\label{app:play_example}

This appendix provides an illustrative example of the pipeline used to construct dialogues in \textsc{PlayDial-FA}. 
For each scene, the original script excerpt is first segmented into a dialogue, then reformulated by \textsc{GPT-4o} to avoid verbatim copying, and finally verified and corrected by human annotators.
\includepdf[pages=-]{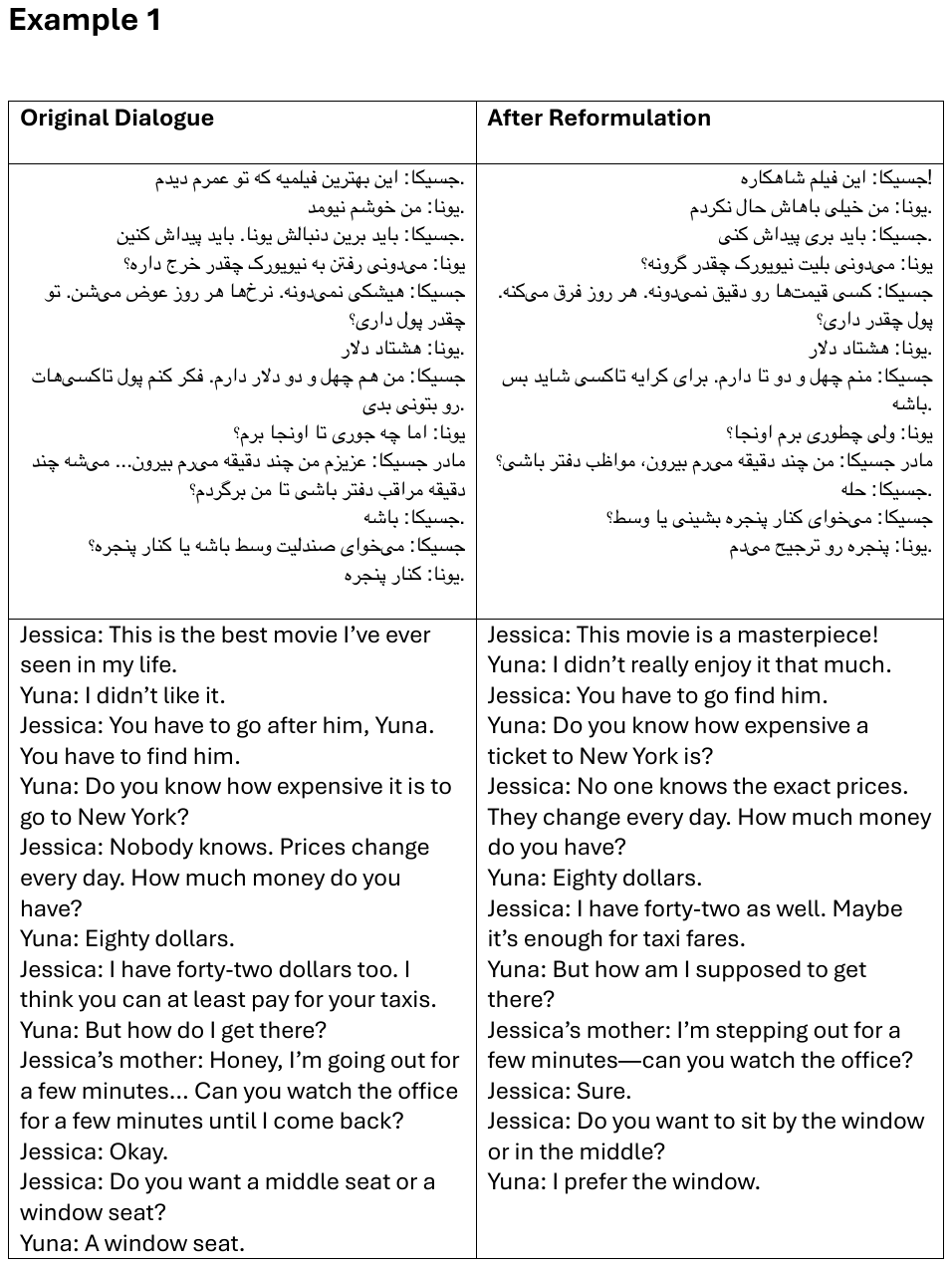}

\section{Human Evaluation Guidelines}
\label{app:human_eval_guidelines}

To assess the quality of generated dialogues in \textsc{Wiki-FaDial}, we conducted a human evaluation study with native Farsi speakers. Annotators evaluated each generated dialogue independently using a five-level quality scale designed to measure conversational naturalness, coherence, fluency, and contextual appropriateness.

Annotators were instructed to read the full dialogue and assign a single overall score according to the following criteria:

\begin{itemize}
    \item \textbf{A } — Highly natural, coherent, fluent, and contextually appropriate dialogue. Responses resemble realistic human conversation with clear conversational flow and culturally natural phrasing.
    
    \item \textbf{B } — Good-quality dialogue with only minor grammatical, fluency, or phrasing issues. The dialogue remains natural, understandable, and contextually consistent.
    
    \item \textbf{C } — Acceptable dialogue with noticeable problems in fluency, coherence, or naturalness, but still understandable overall.
    
    \item \textbf{D } — Weak dialogue containing major inconsistencies, unnatural responses, repetitive phrasing, or poor conversational flow.
    
    \item \textbf{E } — Very poor dialogue that is incoherent, irrelevant, broken, or fails to maintain meaningful conversation.
\end{itemize}

Annotators were instructed to consider the following dimensions during evaluation:

\begin{enumerate}
    \item \textbf{Fluency:} grammatical correctness and readability of individual utterances.
    
    \item \textbf{Coherence:} logical consistency and continuity across dialogue turns.
    
    \item \textbf{Contextual Appropriateness:} relevance of responses to the conversational context.
    
    \item \textbf{Naturalness:} similarity to realistic everyday Farsi conversation.
    
    \item \textbf{Cultural Plausibility:} use of culturally appropriate expressions, politeness conventions, and conversational style.
\end{enumerate}

Annotators were encouraged to evaluate dialogues holistically rather than penalizing isolated minor mistakes. Dialogues with occasional grammatical errors could still receive high scores if the overall interaction remained coherent and natural, while dialogues with strong lexical overlap but unnatural conversational structure were assigned lower scores.

Below, we provide representative examples corresponding to all five evaluation categories (A--E).
\includepdf[pages=-]{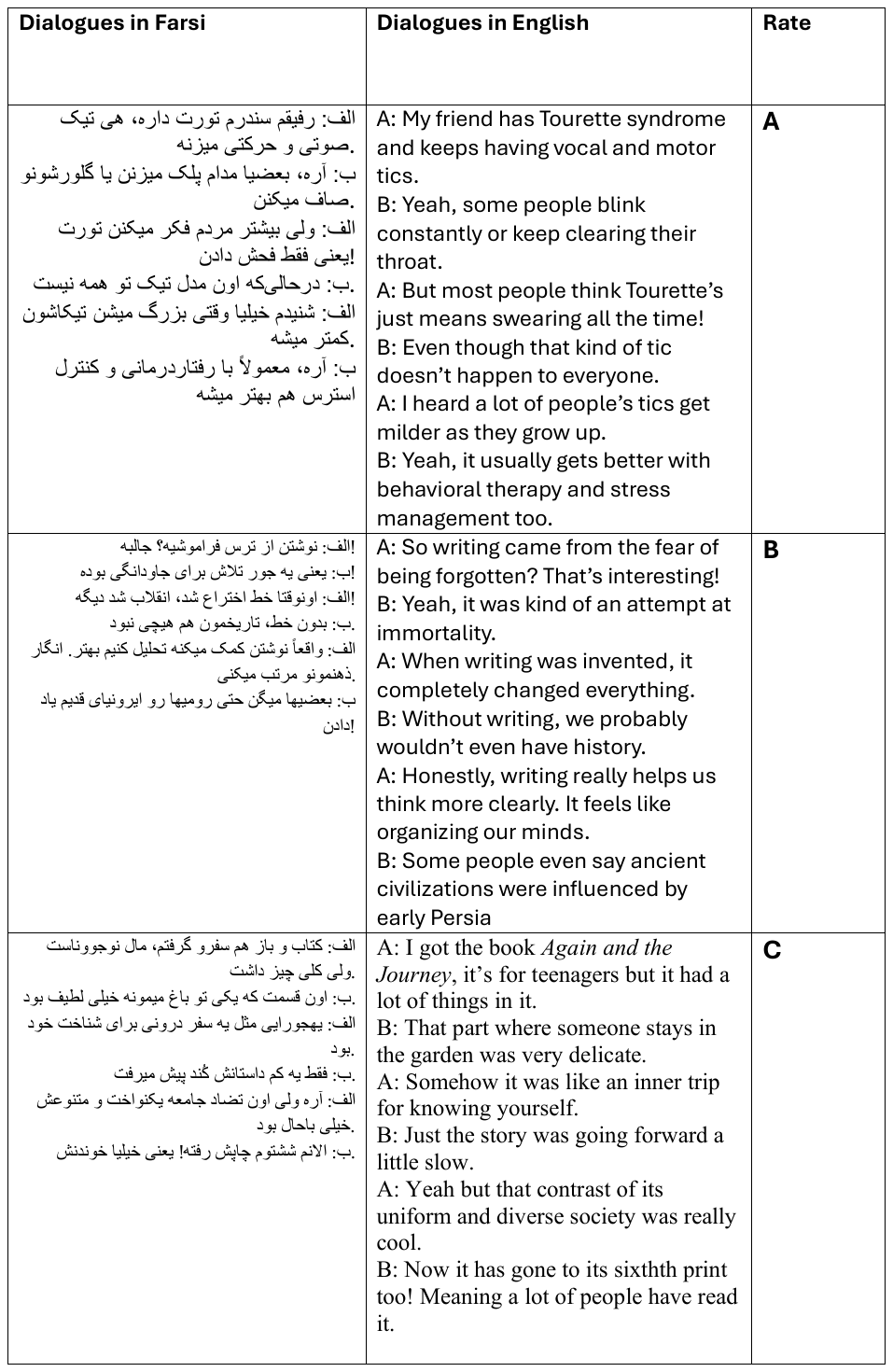}

\section{Independent External Validation}
\label{app:external_validation}

To further assess dataset quality and annotation reliability, we conducted an independent validation study using a native Farsi speaker who was not involved in dataset construction or verification. The annotator evaluated stratified random samples of 100 instances for each annotation task, resulting in 300 evaluation items across the TalkFa datasets.

\subsection{Subjective Quality Evaluation}

The annotator evaluated translation and generation quality using Likert-style scales ranging from 1--3 or 1--4 depending on the evaluation criterion. Table~\ref{tab:external-quality} summarizes the results.

\begin{table*}[!t]
\centering\small
\begin{tabular}{llccc}
\toprule
\textbf{Dataset} & \textbf{Metric} & \textbf{Mean} & \textbf{Scale} & \textbf{Interpretation} \\
\midrule
\multirow{4}{*}{\textsc{Wiki-FaDial}}
& Faithfulness & 3.00 & 3 & Fully faithful to source text \\
& Naturalness & 3.60 & 4 & Highly natural phrasing \\
& Cultural Appropriateness & 2.60 & 3 & Mostly culturally appropriate \\
& LLM Artifacts & 2.60 & 3 & Very few obvious artifacts \\
\midrule
\multirow{3}{*}{\textsc{DailyDialog-FA}}
& Meaning Preservation & 3.00 & 3 & Fully faithful translation \\
& Naturalness & 4.00 & 4 & Fully natural spoken Farsi \\
& Cultural Localization & 2.95 & 3 & Strong localization quality \\
\midrule
\multirow{1}{*}{\textsc{PlayDial-FA}}
& Naturalness & 3.30 & 4 & Natural theatrical dialogue \\
\bottomrule
\end{tabular}
\caption{Independent external evaluation of subjective dialogue quality across the TalkFa datasets.}
\label{tab:external-quality}
\end{table*}

\subsection{Inter-Annotator Agreement}

We further computed raw agreement and Cohen's $\kappa$ between the independent annotator and the original gold annotations.

\begin{table*}[ht]
\centering\small
\begin{tabular}{llcccc}
\toprule
\textbf{Task} & \textbf{Dataset} & \textbf{\# Items} & \textbf{Raw Agreement} & \textbf{$\kappa$} & \textbf{Interpretation} \\
\midrule
Dialogue Acts & \textsc{DailyDialog-FA} & 100 turns & 85\% & 0.800 & Substantial \\
Emotions & \textsc{DailyDialog-FA} & 100 turns & 89\% & 0.872 & Almost Perfect \\
Sentiment & \textsc{PlayDial-FA} & 100 dialogues & 87\% & 0.805 & Almost Perfect \\
\bottomrule
\end{tabular}
\caption{Agreement between the independent annotator and the original gold annotations. Agreement interpretation follows Landis and Koch~\cite{landis1977measurement}.}
\label{tab:external-kappa}
\end{table*}

\subsection{Discussion}

The independent external validation provides additional evidence for the quality and reproducibility of the TalkFa benchmark. Across 300 stratified random samples (100 per annotation task), the independent native Farsi annotator assigned high quality scores to all three datasets. \textsc{Wiki-FaDial} achieved 2.8/3.00 for factual grounding, 3.65/4.00 for naturalness, and 2.70/3.00 for cultural appropriateness. \textsc{DailyDialog-FA} obtained perfect meaning preservation (3.00/3.00), high naturalness (3.90/4.00), and 2.8/3.00 for cultural localization, while \textsc{PlayDial-FA} achieved 3.50/4.00 for dialogue naturalness. These results indicate that the combination of LLM-assisted construction and multi-stage native-speaker verification produces fluent, culturally appropriate dialogues while preserving the intended semantic content.
Annotation reproducibility was likewise consistently high. Dialogue-act annotation achieved 85\% raw agreement with Cohen's $\kappa=0.800$, emotion annotation reached the highest agreement with 89\% ($\kappa=0.872$), and sentiment annotation achieved 87\% ($\kappa=0.805$). Despite the greater subjectivity of emotion and sentiment interpretation, the substantial to almost perfect agreement across all tasks demonstrates that the released annotations are reliable and can be consistently reproduced by an independent annotator.
Overall, the external validation complements the construction-time quality control by confirming both the linguistic quality of the dialogues and the robustness of the released annotations, providing additional evidence for the reliability of the TalkFa benchmark.

\section{Data-Efficiency Ablation}
\label{app:data-efficiency}

To assess data efficiency, we repeat LoRA fine-tuning using \(\{25\%,50\%,75\%,100\%\}\) of the \textsc{Wiki-FaDial} training set and evaluate performance using BERTCos and ROUGE-L.

\begin{figure}[t]
  \centering
  \includegraphics[width=\linewidth]{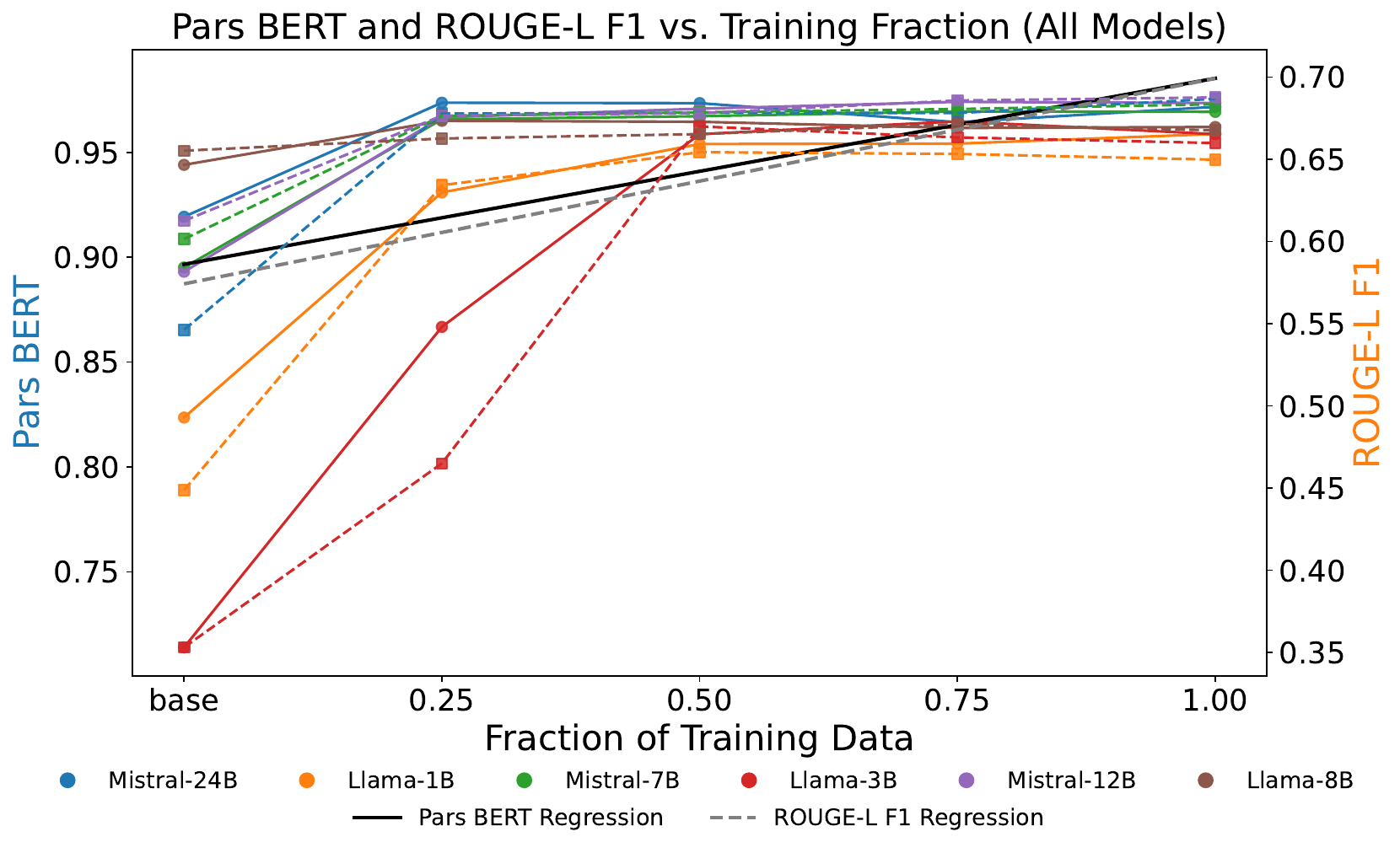}
  \caption{ParsBERT cosine (top) and ROUGE-L F1 (bottom) versus fraction of the training data.}
  \label{fig:wikifadial-ablation}
\end{figure}

Figure~\ref{fig:wikifadial-ablation} shows that performance improves sharply from the Base model to 25\% of the training data, followed by diminishing returns. Across all model sizes, the first quarter of the training set recovers more than 90\% of final performance gains, indicating that relatively small grounded dialogue corpora can effectively teach the task. However, despite these gains, substantial gaps to human-quality dialogue remain, as discussed in Sec.~\ref{sec:beyond-auto}.

\section{Automatic vs.\ Human Evaluation}
\label{app:metric-gap}

Table~\ref{tab:metric-gap} summarizes the disconnect between automatic metrics and human-centered evaluation.

\begin{table*}[ht]
\centering
\small
\begin{tabular}{lcc}
\toprule
\textbf{Metric} & \textbf{Score} & \textbf{Quality Signal?} \\
\midrule
BERTCos (\texttt{bert-fa-base}) & 0.973 & \xmark \\
BERTScore (\texttt{xlm-r})      & 0.930 & \xmark \\
ROUGE-L F\textsubscript{1}      & 0.688 & \xmark \\
\midrule
Human rating (A--E)             & 2.74/5 & \cmark \\
LLM judge (GPT-4.1)             & 3.05/5 & \cmark \\
\bottomrule
\end{tabular}
\caption{Comparison between automatic metrics and human-centered evaluation. \xmark{} indicates that the metric overestimates perceived dialogue quality; \cmark{} indicates a more informative quality signal.}
\label{tab:metric-gap}
\end{table*}

The disparity between high automatic scores ($\geq 0.93$) and moderate human ratings (2.74/5) indicates that semantic overlap metrics alone fail to capture dialogue naturalness, coherence, and contextual appropriateness, making human or LLM-based evaluation necessary for calibrated assessment.

\end{document}